%% file: main.tex
\documentclass[preprint,review,accept,moreauthors,pdftex]{Definitions/mdpi_arxiv} 

\usepackage{comment}
\usepackage{xcolor}
\definecolor{customcolor}{rgb}{1,0,0}
\usepackage{arydshln}
\usepackage{makecell}

\firstpage{1} 
\pubvolume{}
\issuenum{}
\articlenumber{}
\pubyear{}
\copyrightyear{}
\history{}

\Title{Unsupervised Domain Adaptation in Semantic Segmentation: a Review}

\Author{Marco Toldo$^{1}$, Andrea Maracani$^{1}$, Umberto Michieli $^{1}$\orcidA{} and Pietro Zanuttigh$^{1,}$*\orcidB{}}

\AuthorNames{first auth, second auth, Umberto Michieli and Pietro Zanuttigh}

\address[1]{%
$^{1}$ \quad Department of Information Engineering, University of Padova; \{toldomarco, maracani, umberto.michieli, zanuttigh\}@dei.unipd.it}

\corres{Correspondence: zanuttigh@dei.unipd.it}

\abstract{The aim of this paper is to give an overview of the recent advancements in the Unsupervised Domain Adaptation (UDA) of deep networks for semantic segmentation. This task is attracting a wide interest, since semantic segmentation models require a huge amount of labeled data and the lack of data fitting specific requirements is the main limitation in the deployment of these techniques. This problem has been recently explored 
and has rapidly grown with a large number of ad-hoc approaches. This motivates us to build a comprehensive overview of the proposed methodologies and to provide a clear categorization.
In this paper, we start by introducing the problem, its formulation and the various scenarios that can be considered. Then, we introduce the different levels at which adaptation strategies may be applied: namely, at the input (image) level, at the internal features representation and at the output level. Furthermore, we present a detailed overview of the literature in the field, dividing previous methods based on the following (non mutually exclusive) categories: adversarial learning, generative-based,  analysis of the classifier discrepancies, self-teaching, entropy minimization, curriculum learning and multi-task learning. Novel research directions are also briefly introduced to give a hint of interesting open problems in the field. Finally, a comparison of the performance of the various methods in the widely used autonomous driving scenario is presented.}


\begin{document}

\input{sections/introduction.tex}

\input{sections/problem_formulation.tex}
\input{sections/literature_review.tex}

\input{sections/datasets.tex}

\input{sections/conclusion.tex}

\vspace{6pt} 
\funding{Our work was in part supported by the Italian Minister for Education (MIUR) under the ``Departments of Excellence" initiative (Law 232/2016).}


\conflictsofinterest{``The authors declare no conflict of interest''.} 



%

\reftitle{References}


\bibliography{strings,refs}





\end{document}

%% file: sections/introduction.tex
\section{Introduction}
\label{sec:intro}

Over the past few years, deep learning techniques have shown impressive results and have achieved great success in many visual applications.
 However, they typically require a huge amount of labeled data matching the considered scenario to obtain reliable performances. The collection and annotation of large datasets for every new task and domain is extremely expensive, time-consuming and error-prone.
Furthermore, in many scenarios sufficient training data may not be available for various reasons, but it often happens that large amount of data is available for other domains and tasks that are in some way related to the considered one. Hence, the ability of using a model trained on correlated samples from a different task would highly benefit real-world applications for which there is scarce data \cite{wang2018deep}.
These considerations are especially true for semantic segmentation, where the learning architectures require a huge amount of manually labeled data, which is extremely expensive to obtain since a per-pixel labeling is needed.

\subsection{Semantic Segmentation}


\newcommand{\sizefiggg}{0.19}
\begin{figure}[tbp]{}
\setlength\tabcolsep{0.7pt} 
\centering
\begin{tabular}{ccccc}
  \makecell{Color\\Image} & Classification & \makecell{Object\\Localization} & Segmentation & \makecell{Semantic\\Segmentation} \\

   \includegraphics[width=\sizefiggg\linewidth]{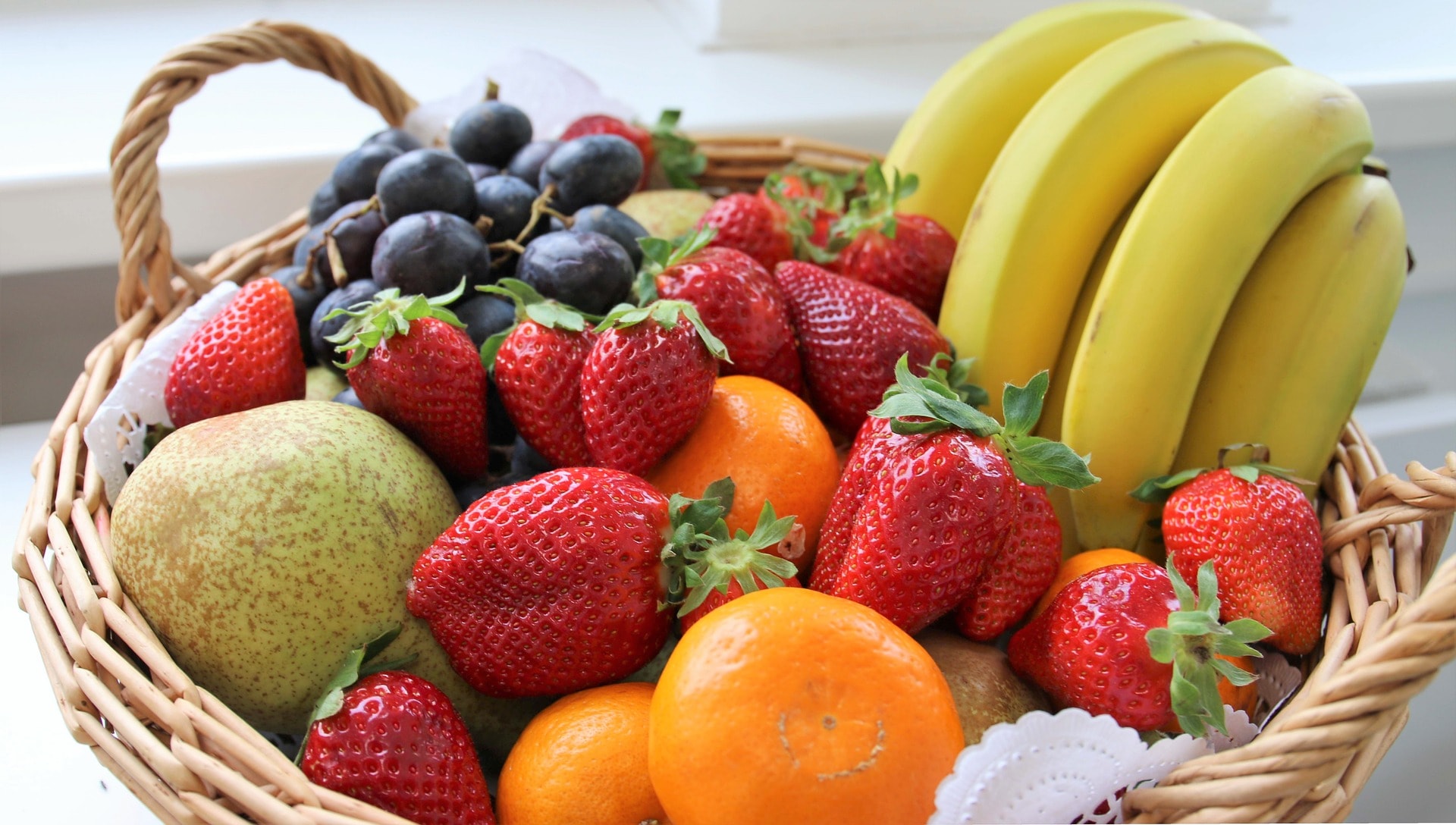} &
   \includegraphics[width=\sizefiggg\linewidth]{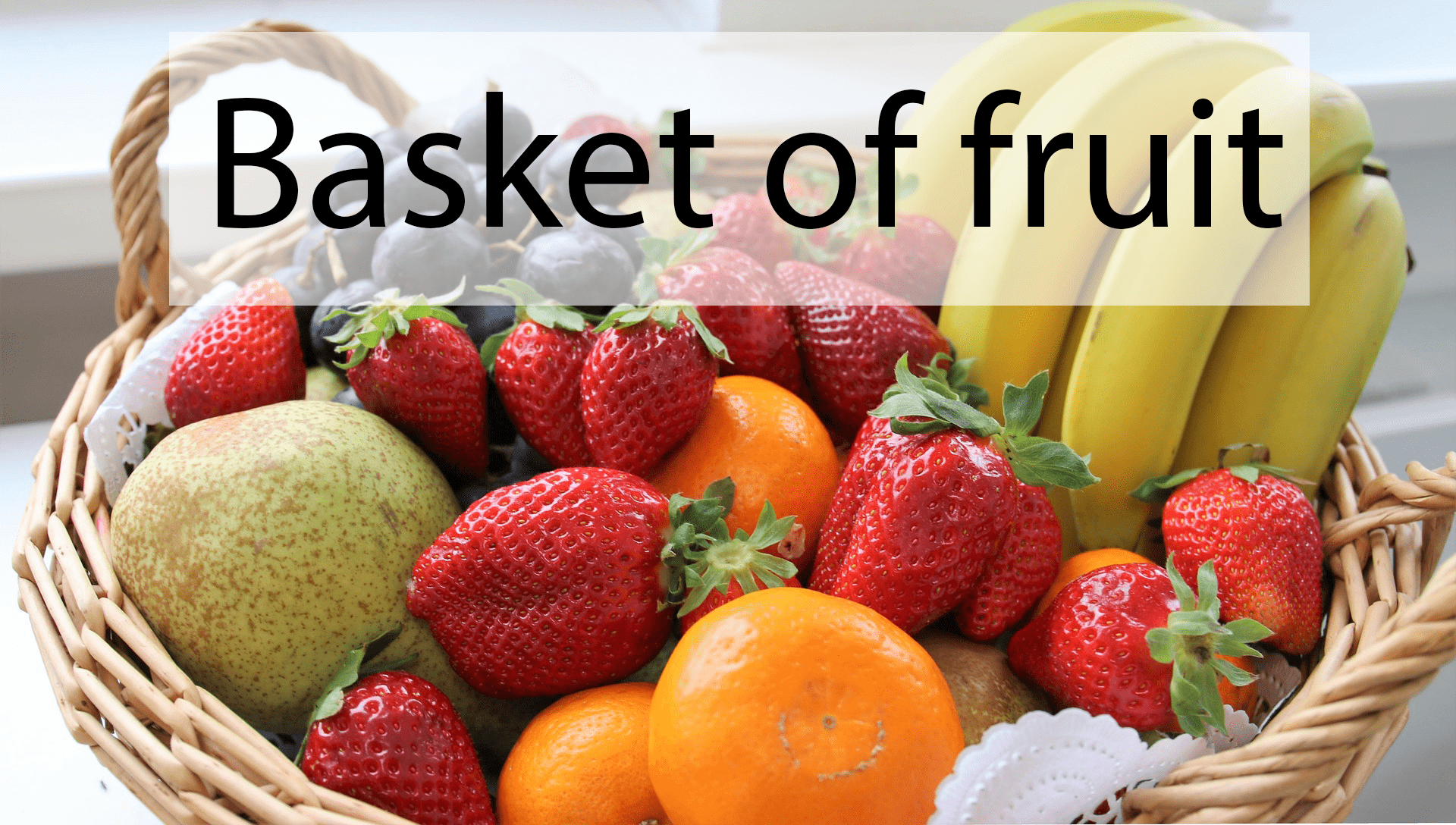} &
   \includegraphics[width=\sizefiggg\linewidth]{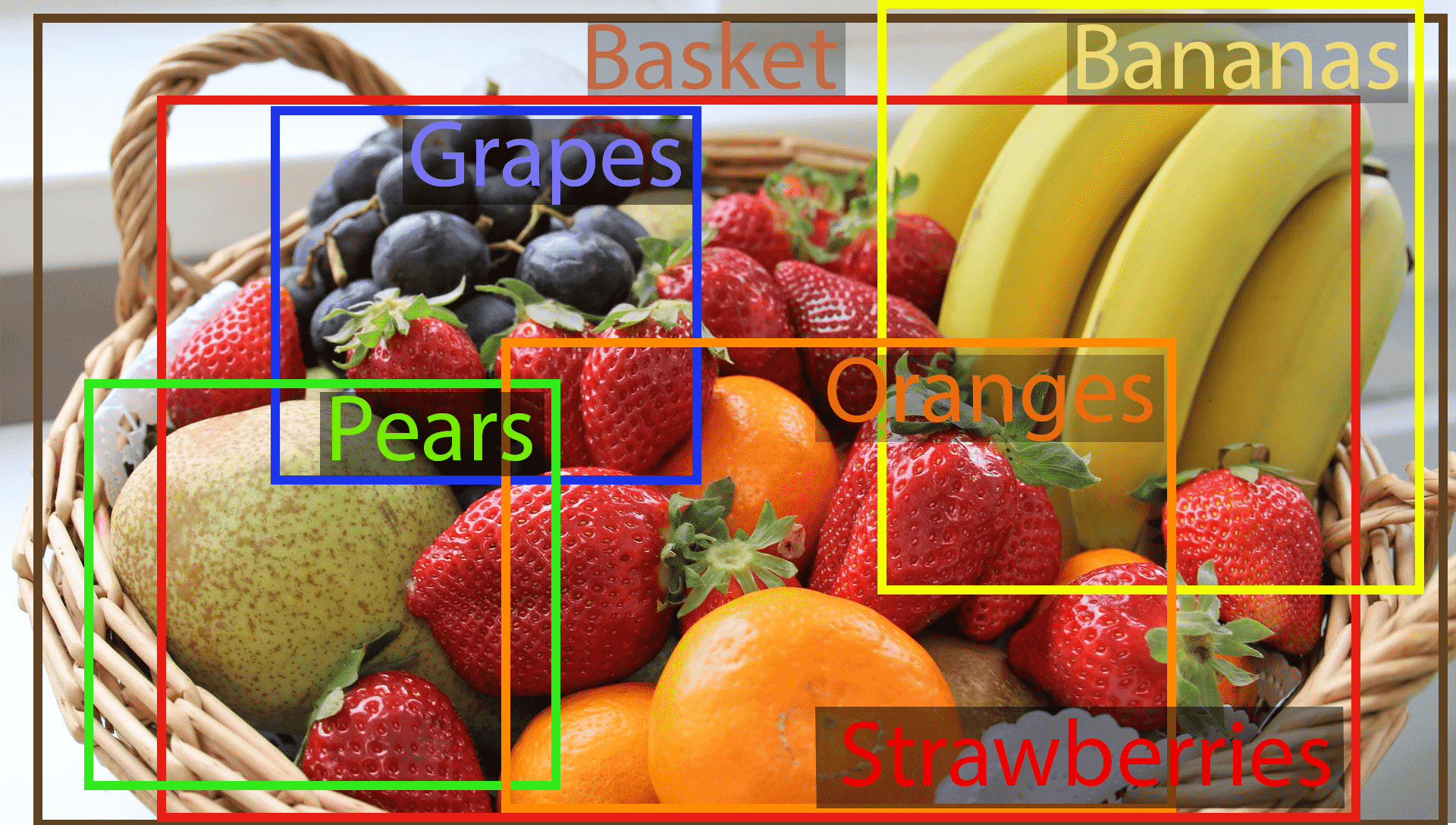} &
   \includegraphics[width=\sizefiggg\linewidth]{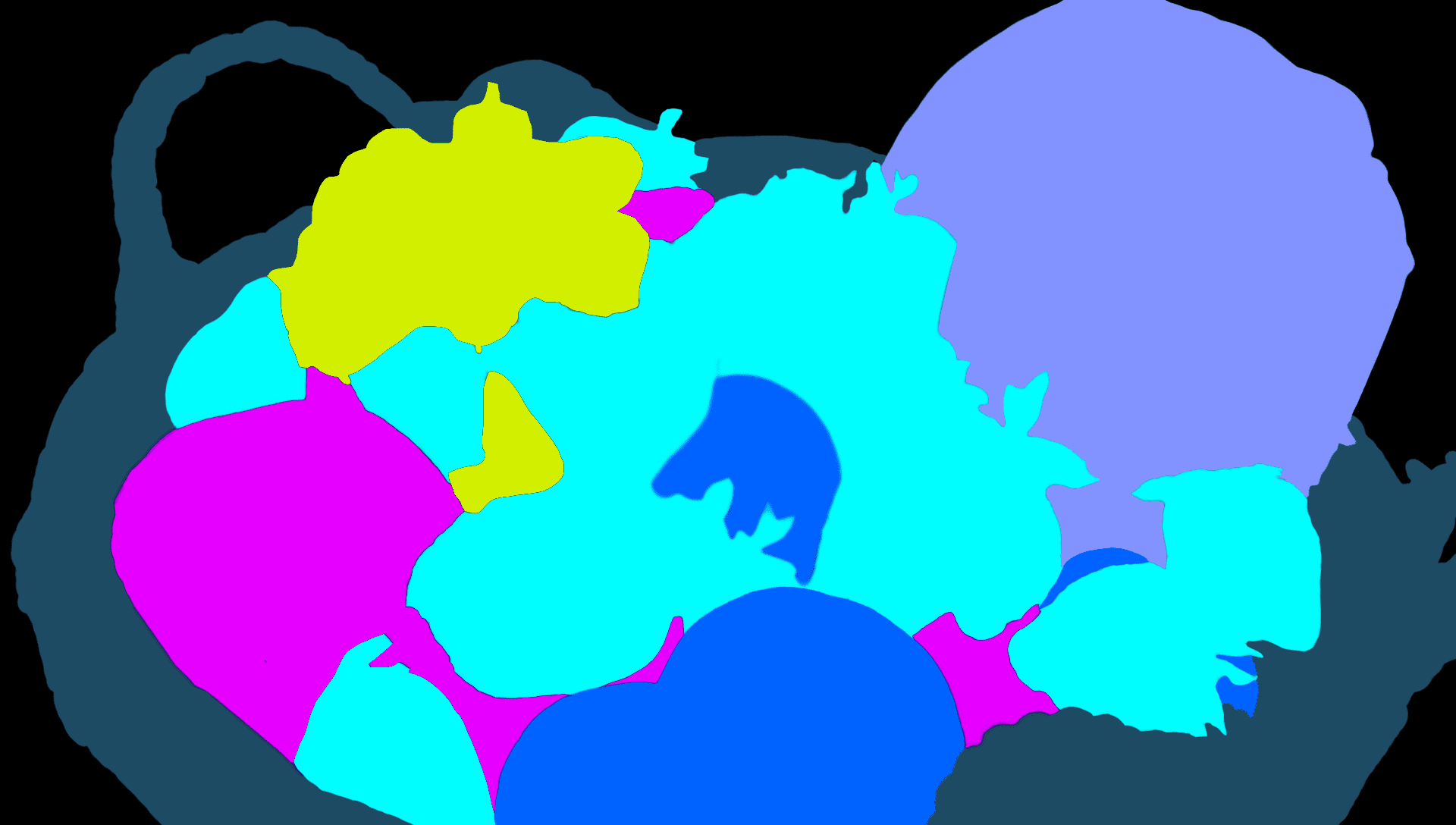} &
   \includegraphics[width=\sizefiggg\linewidth]{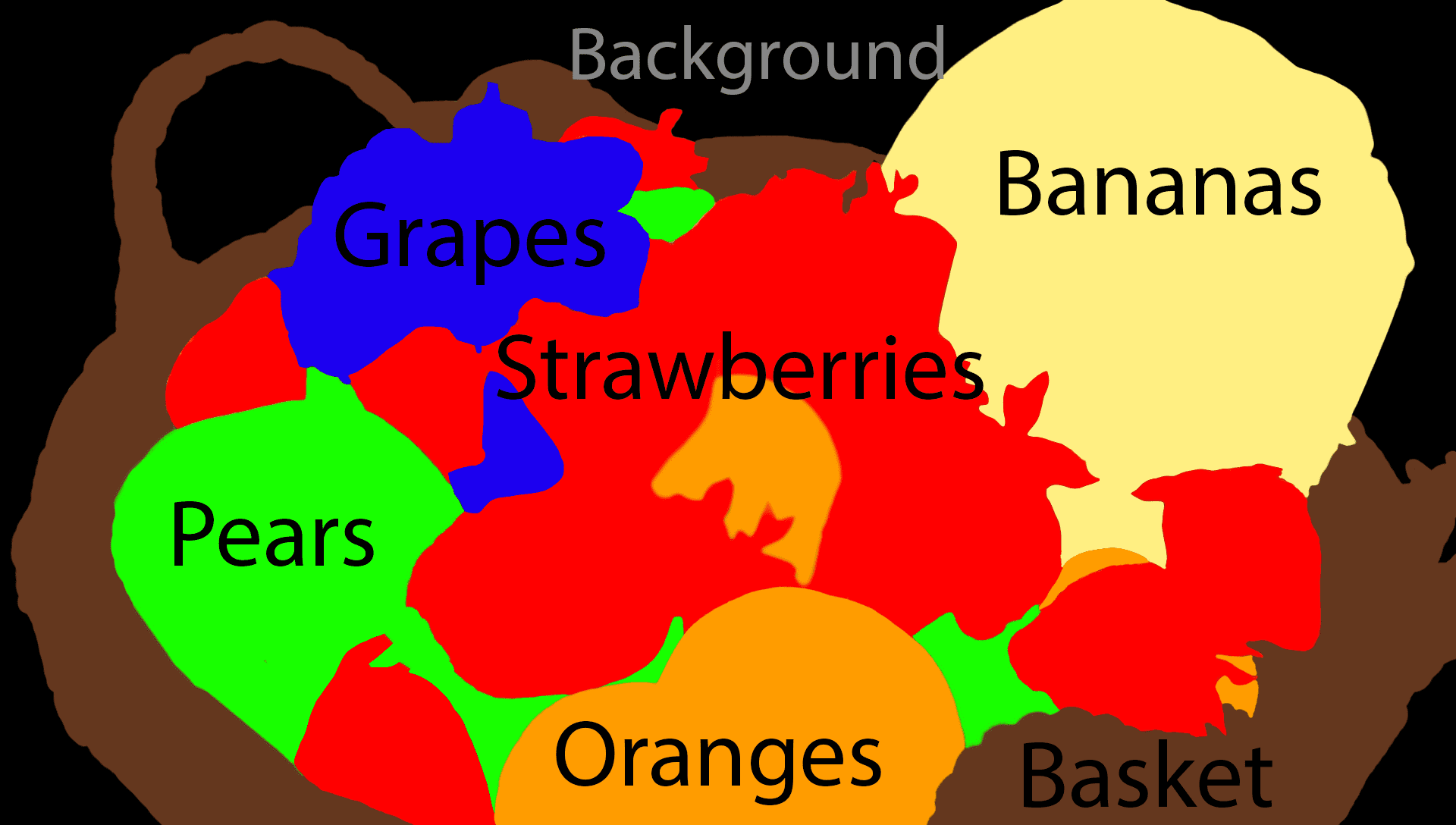} \\
   
   \includegraphics[width=\sizefiggg\linewidth]{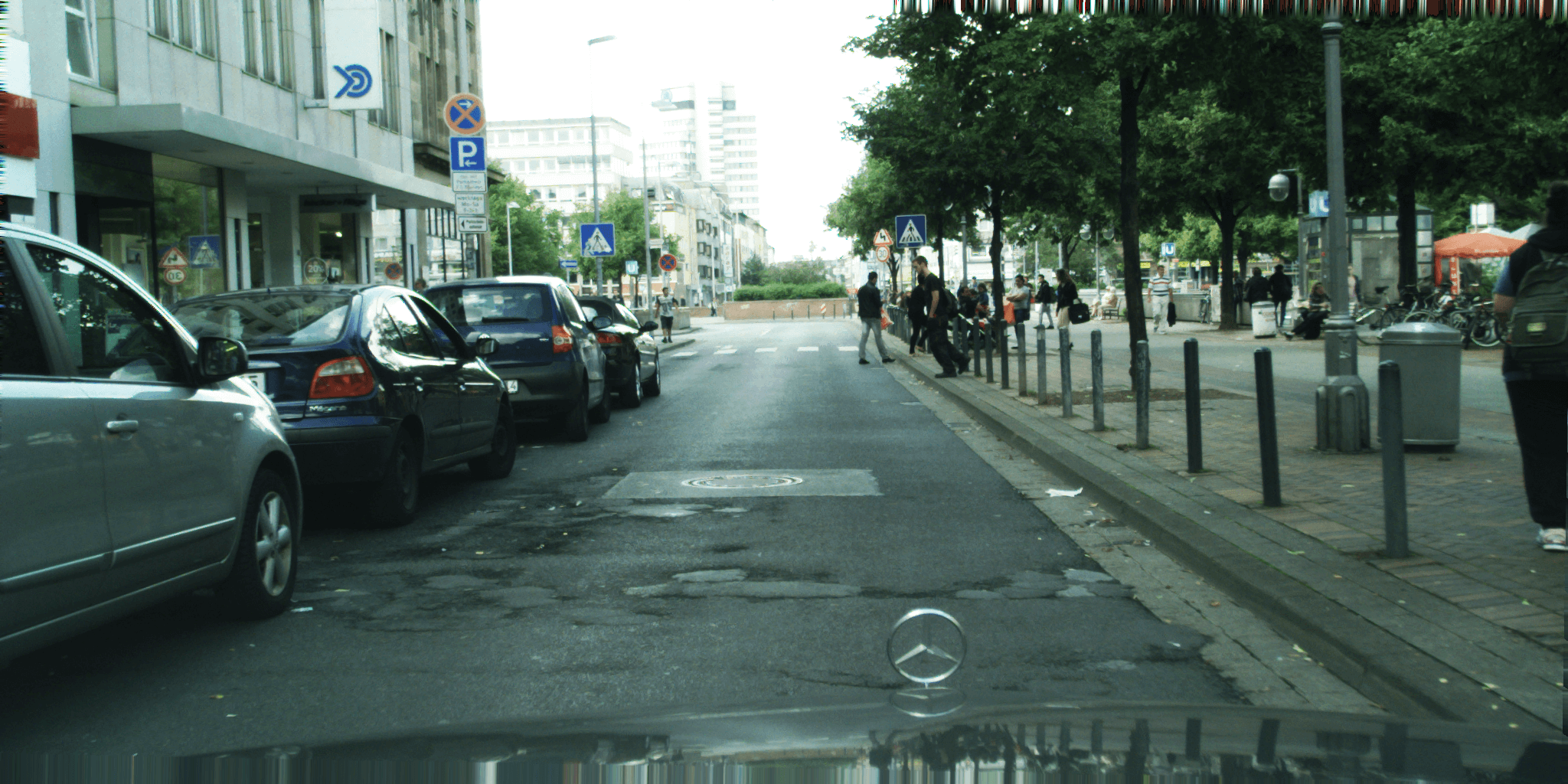} &
   \includegraphics[width=\sizefiggg\linewidth]{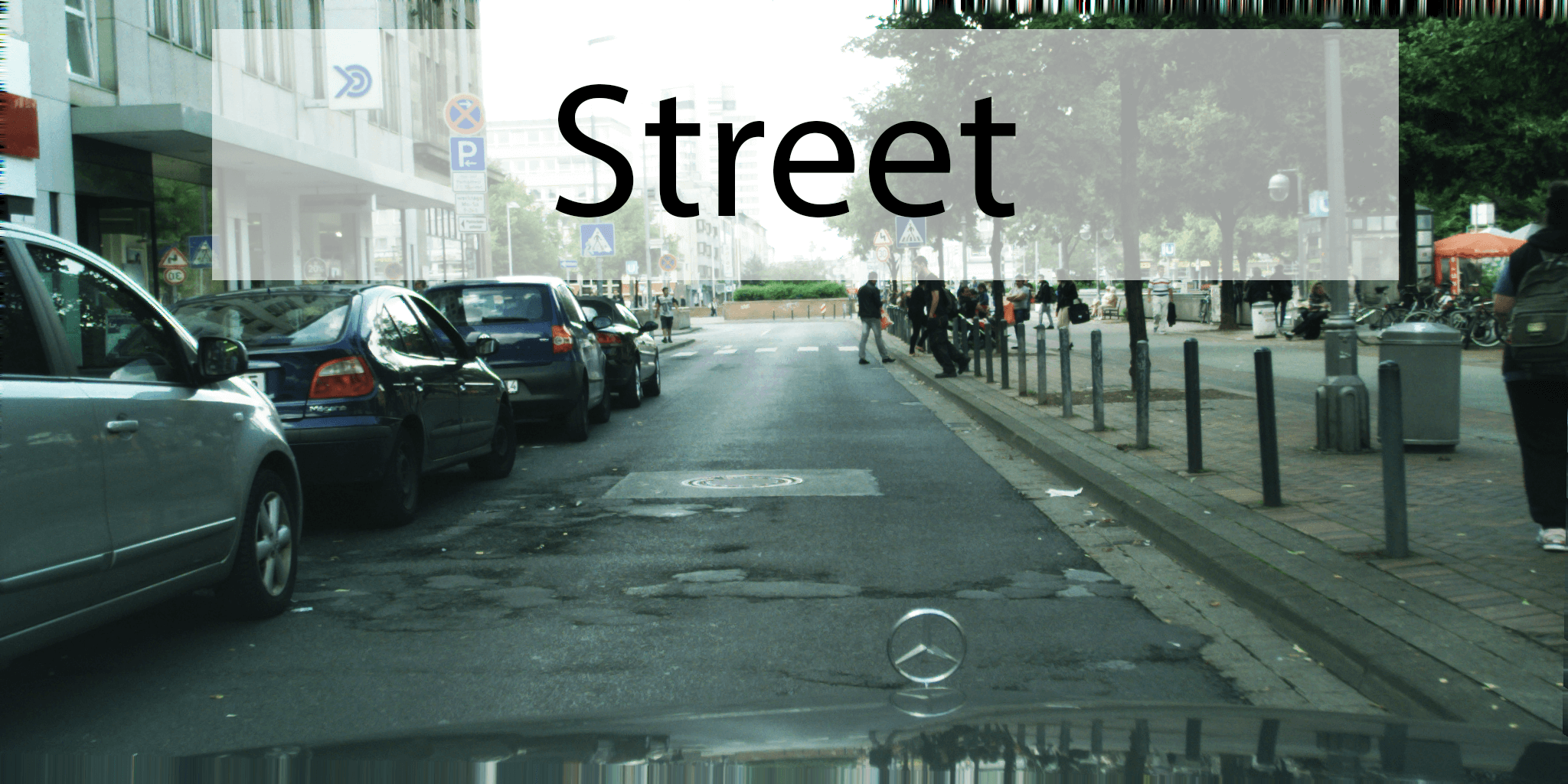} &
   \includegraphics[width=\sizefiggg\linewidth]{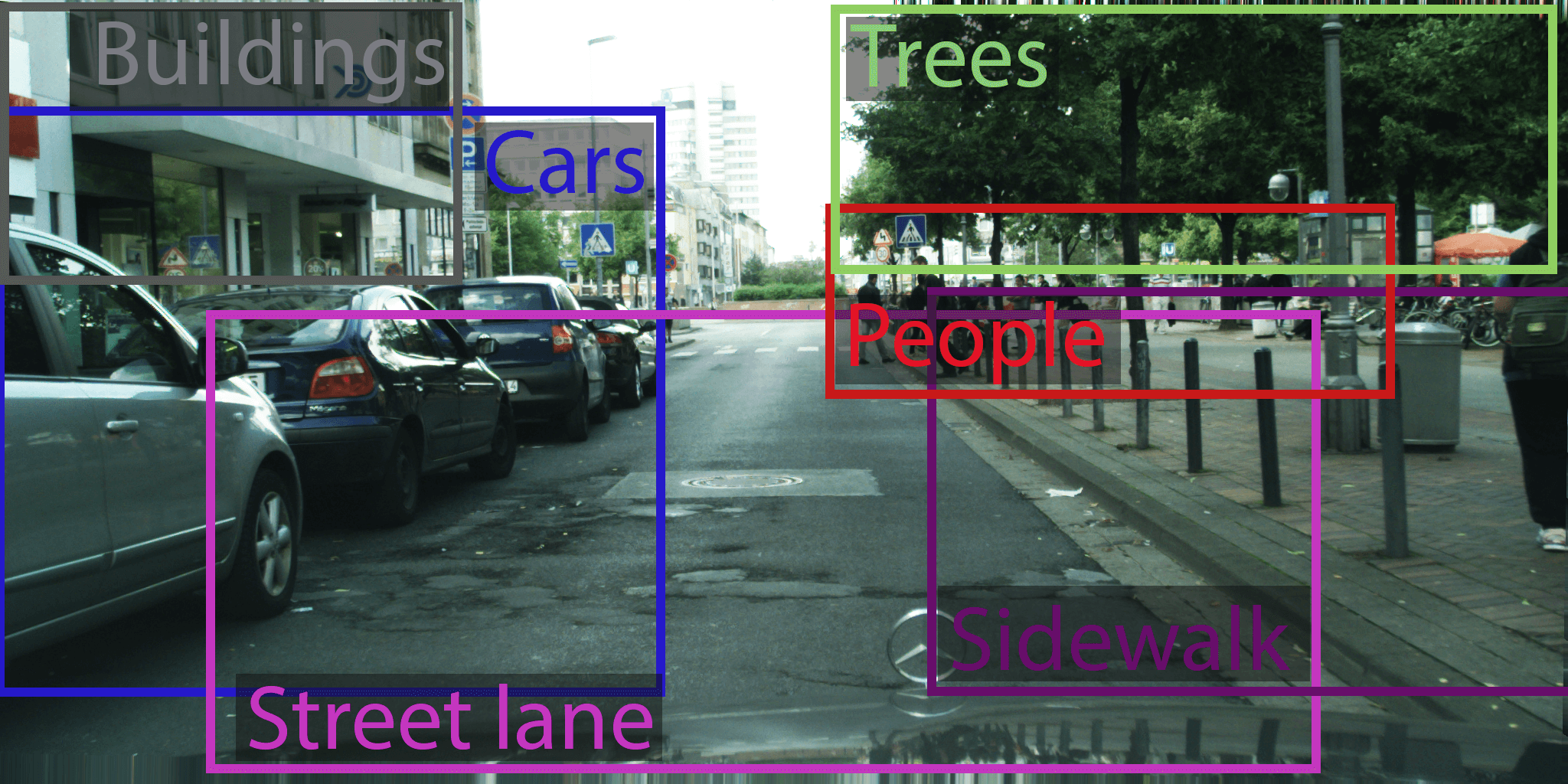} &
   \includegraphics[width=\sizefiggg\linewidth]{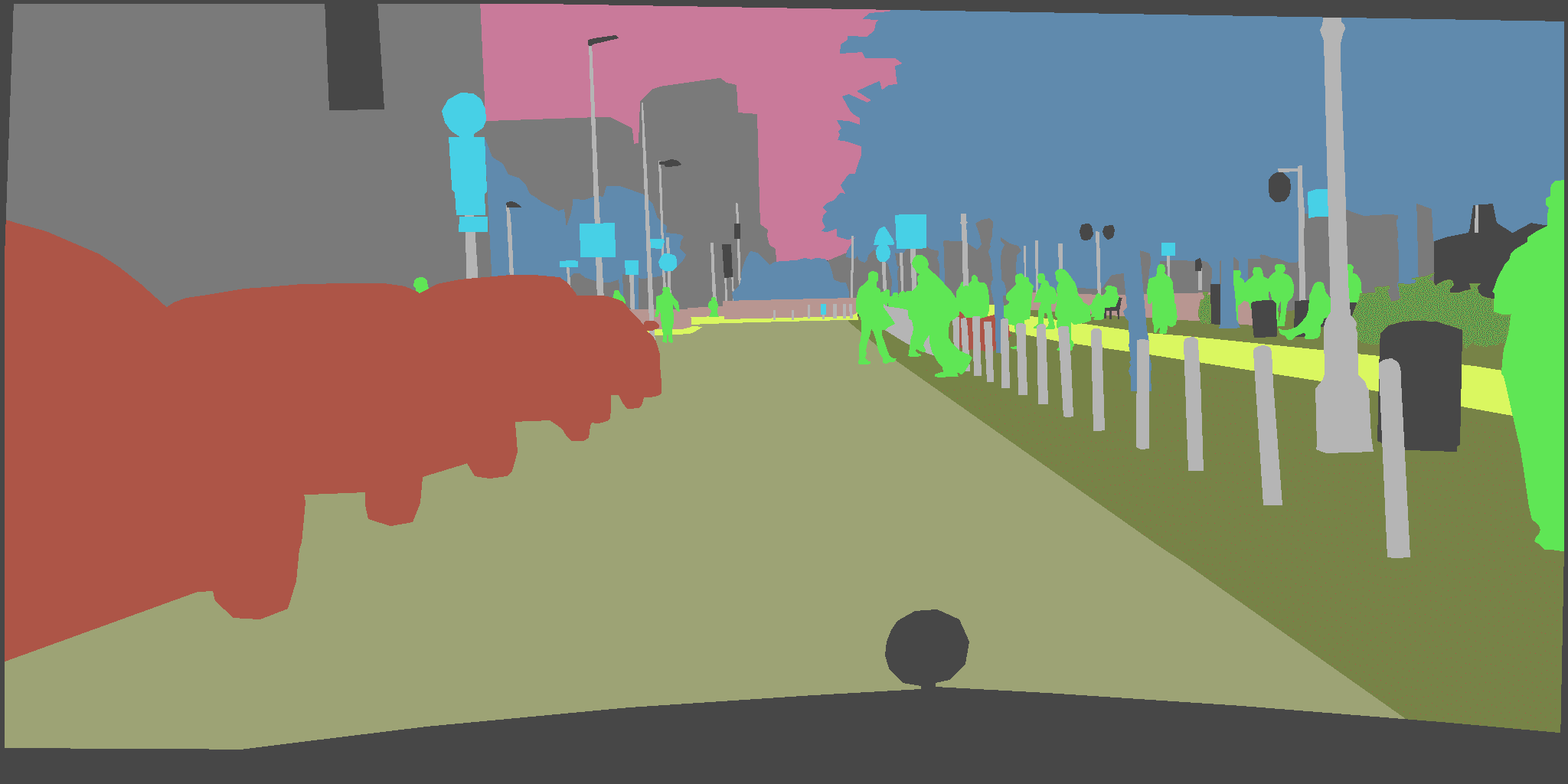} &
   \includegraphics[width=\sizefiggg\linewidth]{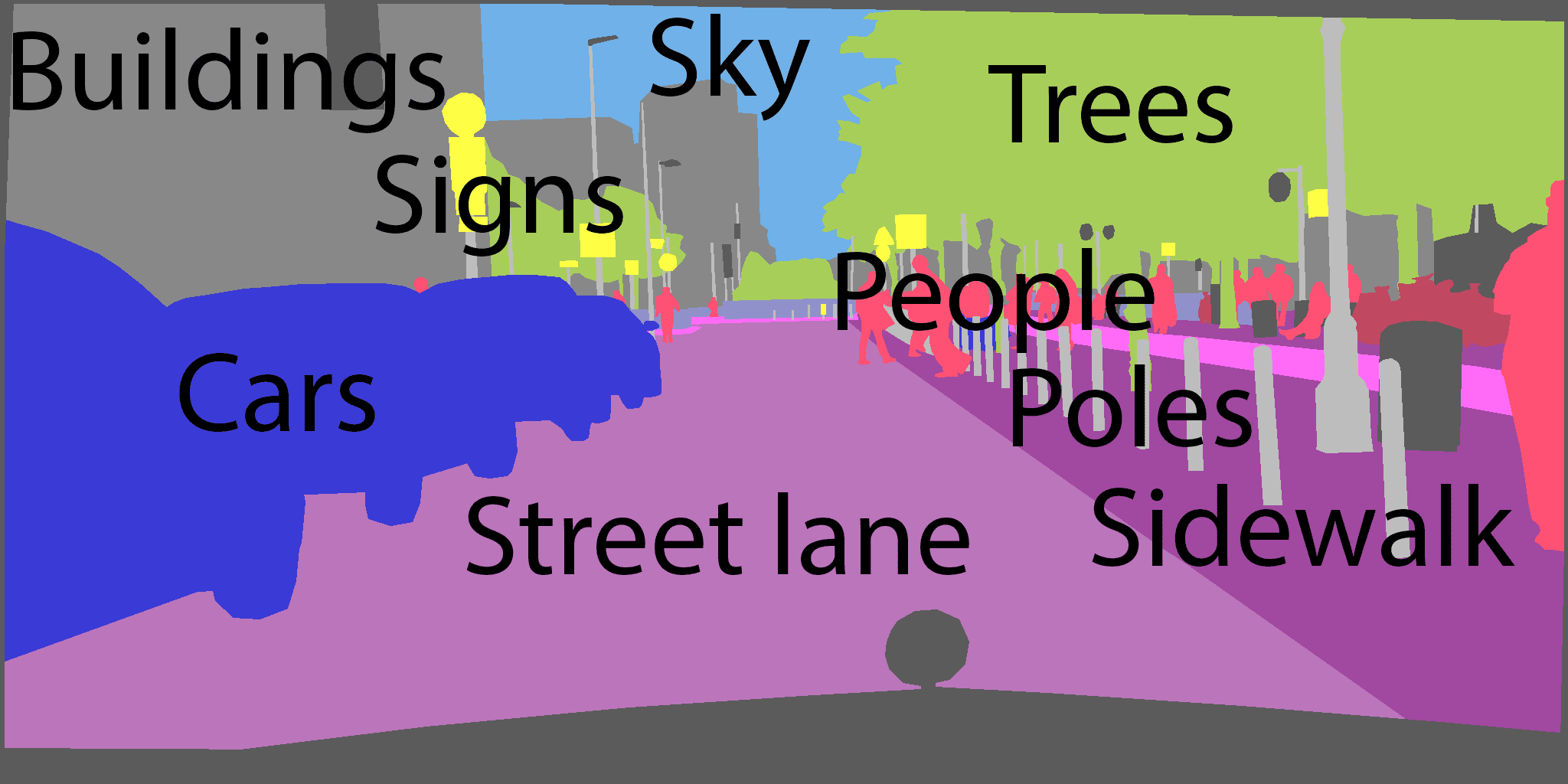} \\
   
   \includegraphics[width=\sizefiggg\linewidth]{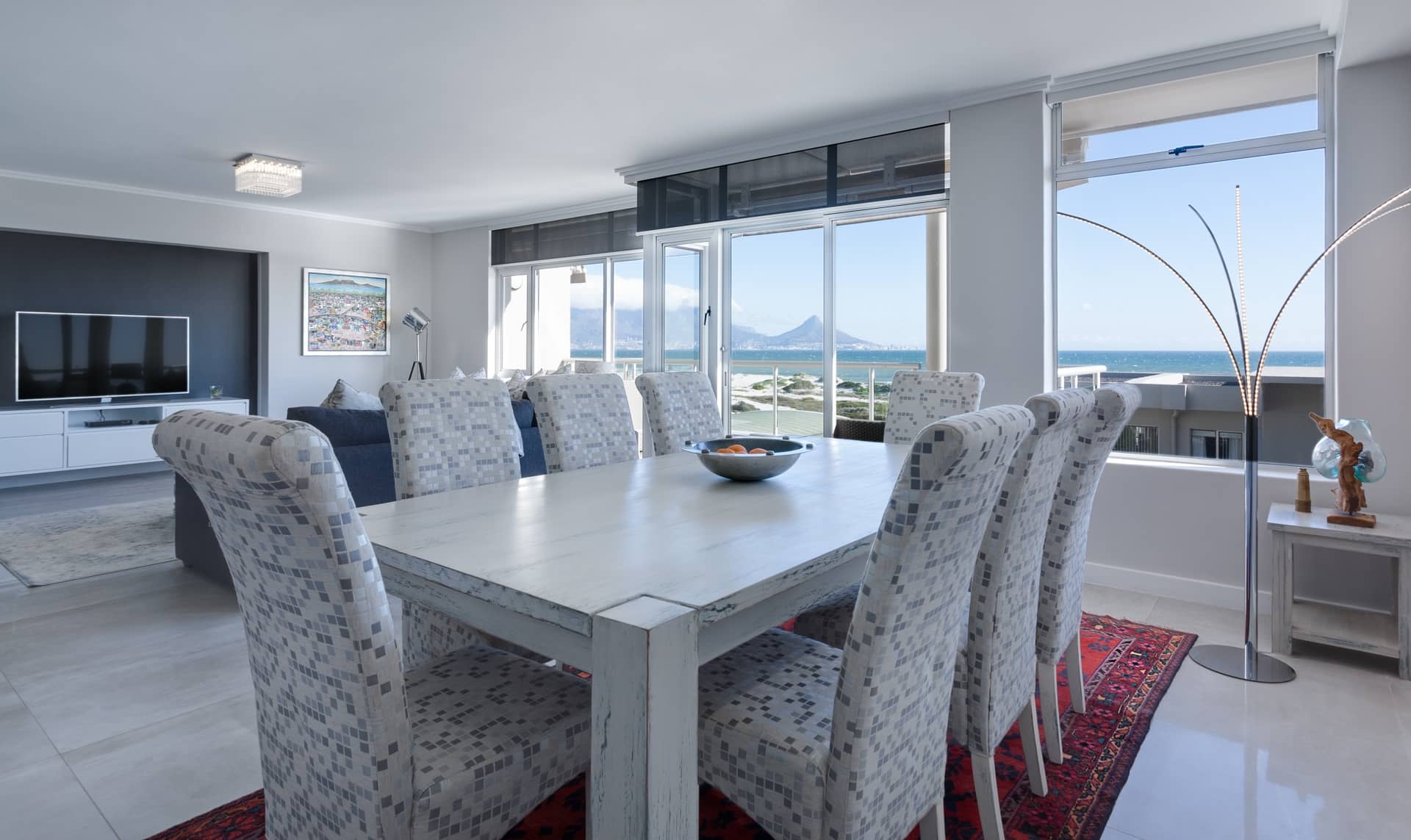} &
   \includegraphics[width=\sizefiggg\linewidth]{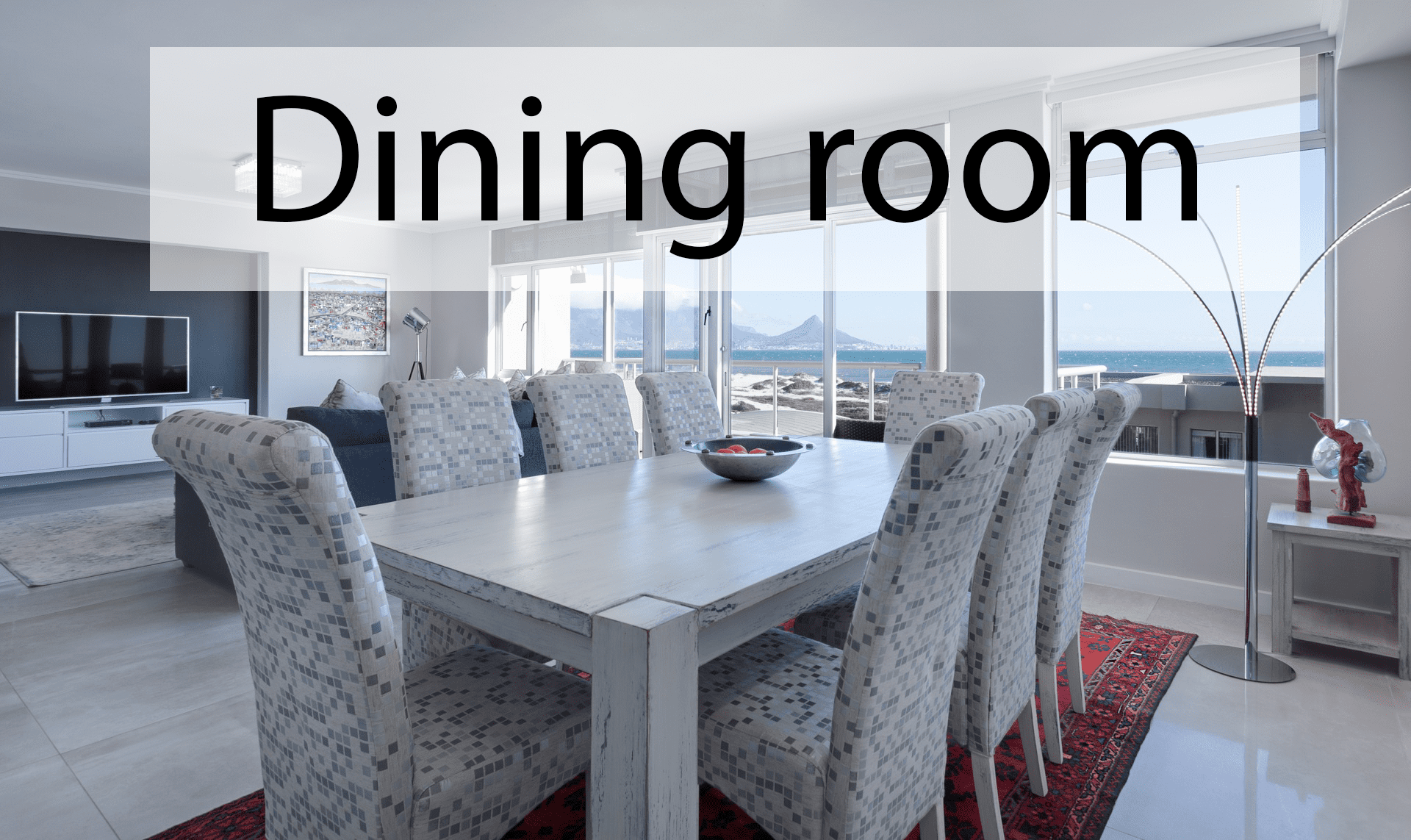} &
   \includegraphics[width=\sizefiggg\linewidth]{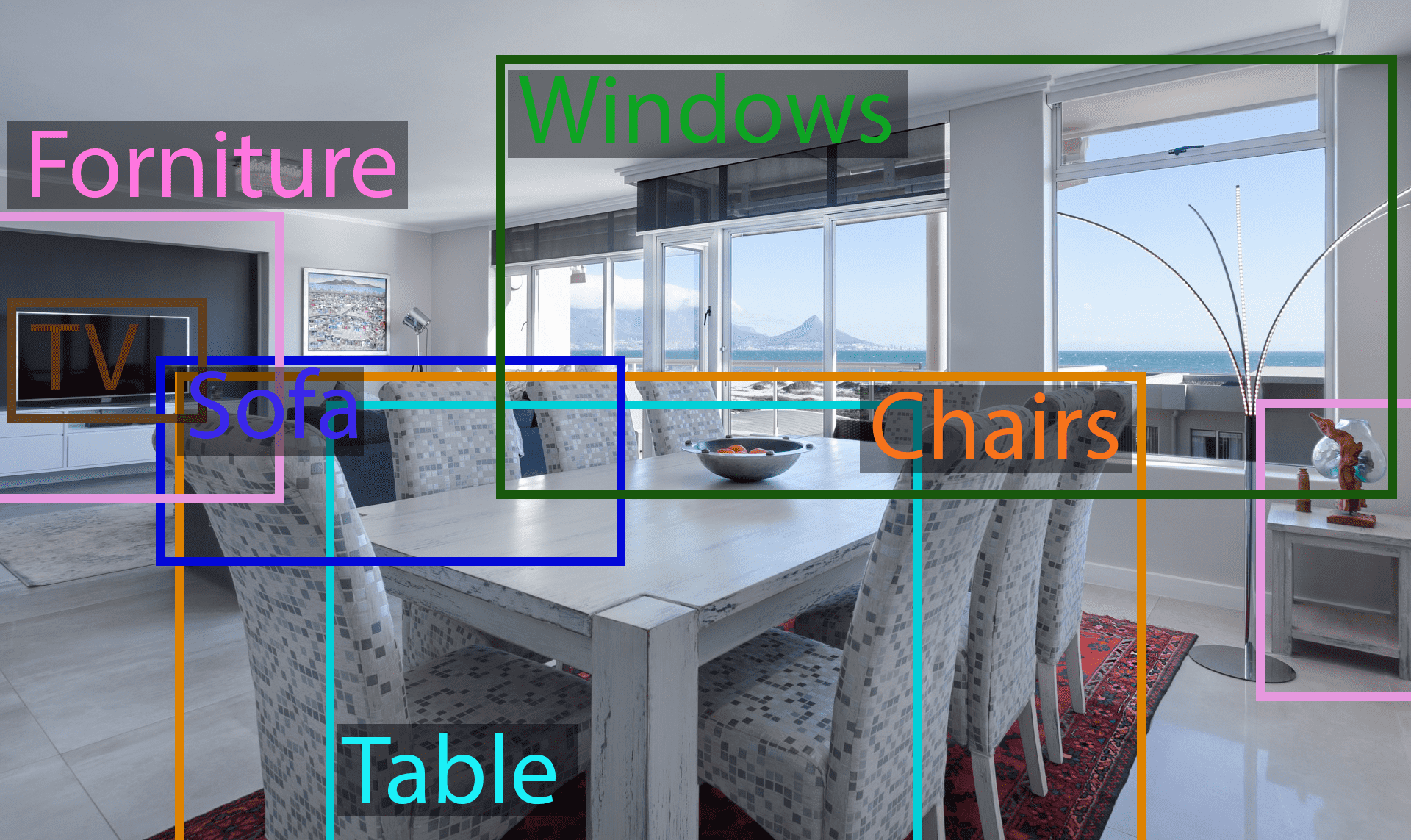} &
   \includegraphics[width=\sizefiggg\linewidth]{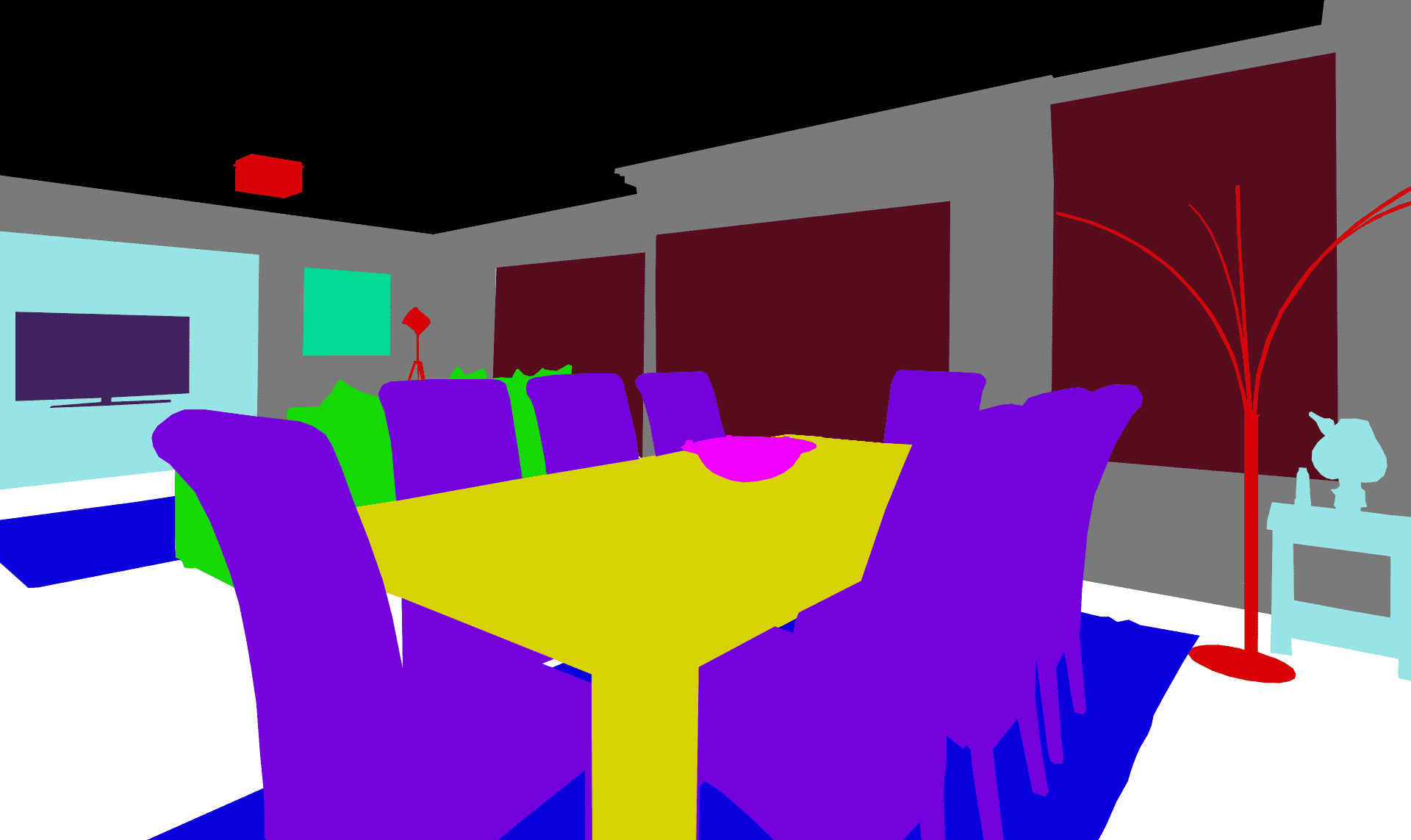} &
   \includegraphics[width=\sizefiggg\linewidth]{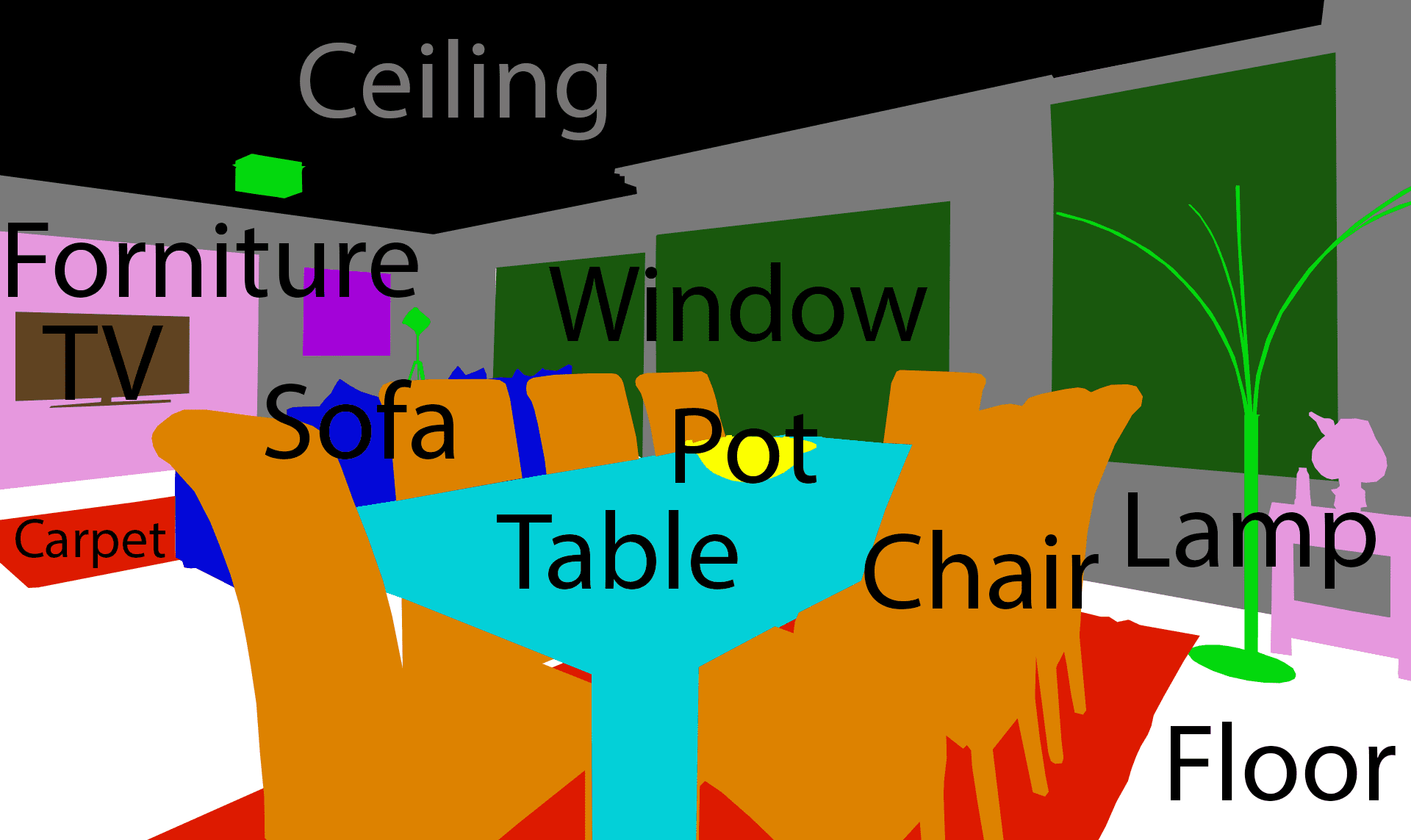} \\

 \end{tabular}
\vspace{-0.1cm}
\caption{Overview of some possible visual tasks on a few sample images from classification (sparse task) to semantic segmentation (dense task).}
\label{fig:tasks}
\end{figure}

Semantic segmentation is one of the most challenging tasks in automatic visual understanding, leading to a deeper understanding of the image content if compared with simpler problems like image classification or object detection. An overview of the most common visual tasks is given in Figure~\ref{fig:tasks}.
In \textit{image classification}, a single label is assigned to the whole image and denotes the pre-dominant object in the scene. In \textit{object localization}, the objects are identified by means of a bounding box and a label is assigned to each box. In \textit{image segmentation}, the scene is clustered into regions corresponding to the various objects and structures but the regions are not labeled. 
\textit{Semantic segmentation}, instead, is the task of assigning to each pixel in the image a label corresponding to its semantic content. For this reason, it is often referred to as a dense labeling task as opposite to other simpler problems where fewer labels are given as output.
Semantic segmentation is a very wide research field and a huge number of approaches have been proposed to tackle it. In particular, deep learning architectures have recently allowed to obtain substantial improvements.

Historically, semantic segmentation has moved its origins as an enriched representation and understanding of the scene with respect to the simpler task of image classification: 
the advent of novel problems to address requiring an higher level of interpretation of the scenes and the possibility to accomplish it, thanks to novel architectures and paradigms (e.g., deep learning), have paved the way to the wide success of semantic image segmentation.

While image classification allows to classify what is contained in an image at a macroscopic level (i.e., one label is assigned to each image), semantic image segmentation generates a pixel-wise mask of each object in the images (i.e., one label is assigned to each pixel of each image). Being the former a much simpler task, it has been tackled since long time with both traditional techniques (such as SVM, LDA,...) and, more recently, with deep learning ones. For this reason, some early-stage works in semantic segmentation build up from classification works, adapting and extending them. 
The most recent state-of-the-art approaches rely on an autoencoder structure, composed by an encoder and a decoder in order  to extract global semantic clues, while retaining input spatial dimensionality. 

Starting from the well-known Fully Convoultional Networks (FCN) architecture \cite{long2015fully}, many models have been proposed, such as PSPNet \cite{zhao2017pyramid}, DRN \cite{yu2017drn} and the various versions of the DeepLab architecture \cite{chen2018deeplab,chen2018encoder,chen2017rethinking}.
These models can achieve impressive performance, but this is strictly related to the availability of a massive amount of labeled data required for their training.
For this reason, even though the pixel-wise annotation procedure is highly expensive and time consuming, many datasets have been created: for example, Cityscapes \cite{cordts2016cityscapes} and Mapillary \cite{neuhold2017mapillary} for
urban scenes; Pascal VOC \cite{everingham2010pascal}, MS-COCO \cite{lin2014microsoft} and ADE20K \cite{zhou2017scene} for visual objects in common contexts; NYUD-v2\cite{silberman2012indoor} and SUN-RGBD \cite{song2015sun} for indoor scenes with depth information. 
In light of these considerations, many recent works try to exploit knowledge extracted from other sources or domains, where labels are plentiful and easily accessible, to reduce the amount of required manually annotated data. 

\subsection{Domain Adaptation (DA)}

Most machine learning models, including Neural Networks (NNs), typically assume that training and test samples are drawn according to the same distribution. However, there are many cases in practical problems where the training and the test data distributions differ. In this survey we focus on the case where a model is trained in one or more domains (called source domains) and then applied  in another different, but related, domain (called target domain) \cite{sun2015survey}. Such learning task is known as Domain Adaptation (DA) and is a fundamental problem in machine learning. Nowadays, it has gained a wide attention from the scientific community and represents a long-standing problem in many real-world applications, such as computer vision \cite{csurka2017domain}, natural language processing \cite{jiang2007instance}, sentiment analysis \cite{fang2014domain}, email filtering, and several others.

Domain Adaptation can be regarded as a particular case of Transfer Learning (TL) that utilizes labeled data in one or more relevant source domains to execute new tasks in a target domain. The aim of DA methodologies is to address the distribution change or the domain shift, which typically greatly degrades the performance of the models \cite{jiang2008literature}. Over the past decades, various DA methods have been proposed to address the shifts between the source and target domains for both traditional machine learning strategies and recent deep learning architectures. 
The intrinsic nature of source and target domains highly influence the final performance of the DA algorithms. Indeed, they are assumed to be somehow related to each other, but not identical. The more correlated they are, the easier the DA task becomes, allowing to achieve high results on the test data. Hence, a key ingredient for a well-performing strategy is the ability of discovering suitable source data to extract useful clues from.

\subsection{Unsupervised Domain Adaptation (UDA)}
The domain adaptation task can be performed using only data from the source domain or using also some samples from the target domain. The simplest solution that could be adopted is to train only on labeled samples from the source domain without using data from the target domain, hoping that no adaptation is needed (source only). In practice, this leads to poor performances, even when only a small visual domain shift exists. To cope with this, UDA approaches exploit labeled samples from the source domain and unlabeled samples from the target one (source to target UDA).

Especially in the semantic segmentation task where a pixel-by-pixel labeling is required, the samples annotation is the most demanding task, while data acquisition is much simpler and cheaper. For this reason, in this survey we will cover the scenario that takes the name of \textit{Unsupervised Domain Adaptation (UDA)}. Indeed, it is the most interesting in our specific setting since there is no direct supervision on the target domain (i.e., no labels of the target domain are required).\\
In this scenario, the typical assumption is that the source and target domains are different but in some way related (e.g., the source could be synthetically generated data resembling the real world representations in the target one). Typically, an initial supervised training on the source domain is adapted to the target one by means of various unsupervised learning strategies aiming at achieving good performance also on the target domain (for which no labels are available). In the standard setting, the set of target classes are the same, but advanced settings where also the target labels change can be considered (see Section \ref{sec:subsec:problem}).

\subsection{Application Motivations}

There exists a large number of applications which may significantly benefit from UDA.
In general, each application focuses on a very peculiar setting with images taken with a specific camera and a particular environment to solve a prefixed task. The first and easiest solution is to get as much labeled data as possible for the specific problem, but, as already mentioned, this is unfortunately very time consuming and expensive, thus making it unfeasible in many real-world contexts.
On the other hand, large and publicly available labeled datasets typically contain generic data and their direct use in specific applications does not grant good performance in the relevant application-specific domain. 
A second solution would be to transfer source knowledge acquired on a broader scenario and adapt it to the specific setup being targeted. Such context, for example, is fairly common in industrial applications.

An example application is face recognition, which represents a challenging problem that has been actively researched for many years. 
Current models for face recognition perform very well when training and testing images are acquired under controlled conditions. However, their accuracy quickly degrades when the test images contain variations that are not present in the training images \cite{patel2015visual}. For instance, these variations could be change in pose, illumination or view point, and depending on the composition of training and test sets this can be regarded as a domain adaptation problem \cite{ho2014model,patel2015visual}.

Another straightforward application lies in object recognition, where one may be interested in adapting object detection capabilities from a typically larger set to a specific small-size dataset \cite{saenko2010adapting}.

Furthermore, the recent improvements in the computer graphics field allowed the production of a large amount of synthetic data for many vision-related tasks. This allows to easily obtain large training sets but on the other side the domain shift between synthetic and real world data needs to be addressed. 
In this field, the most predominant application is found in autonomous vehicles scenarios as will be further discussed in Section \ref{sec:datasets}.

\begin{figure}[t]{}
\setlength\tabcolsep{0.7pt} 
\centering
\begin{tabular}{ccc}
   \includegraphics[width=0.32\linewidth]{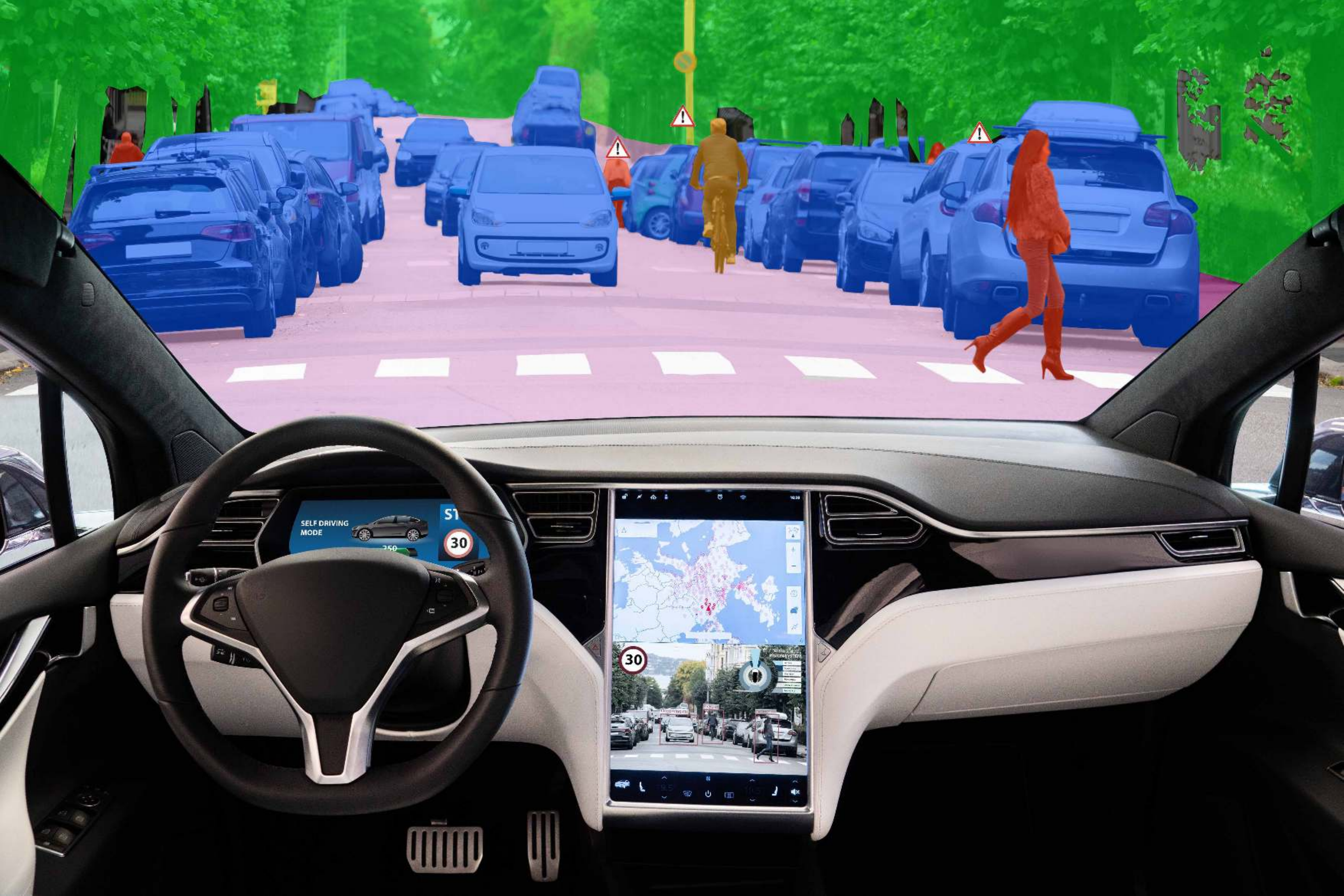} &
   \includegraphics[width=0.32\linewidth]{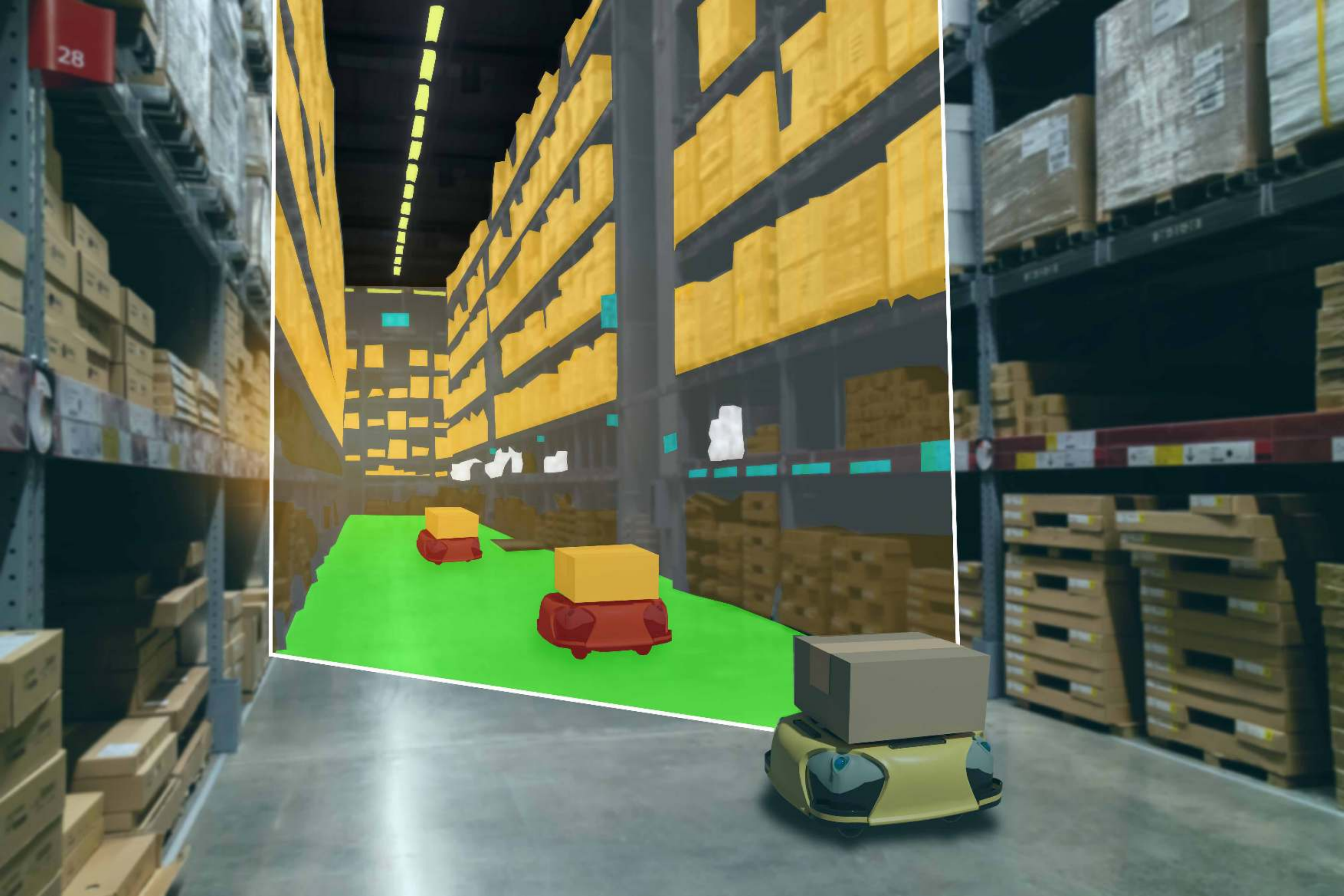} &
   \includegraphics[width=0.32\linewidth]{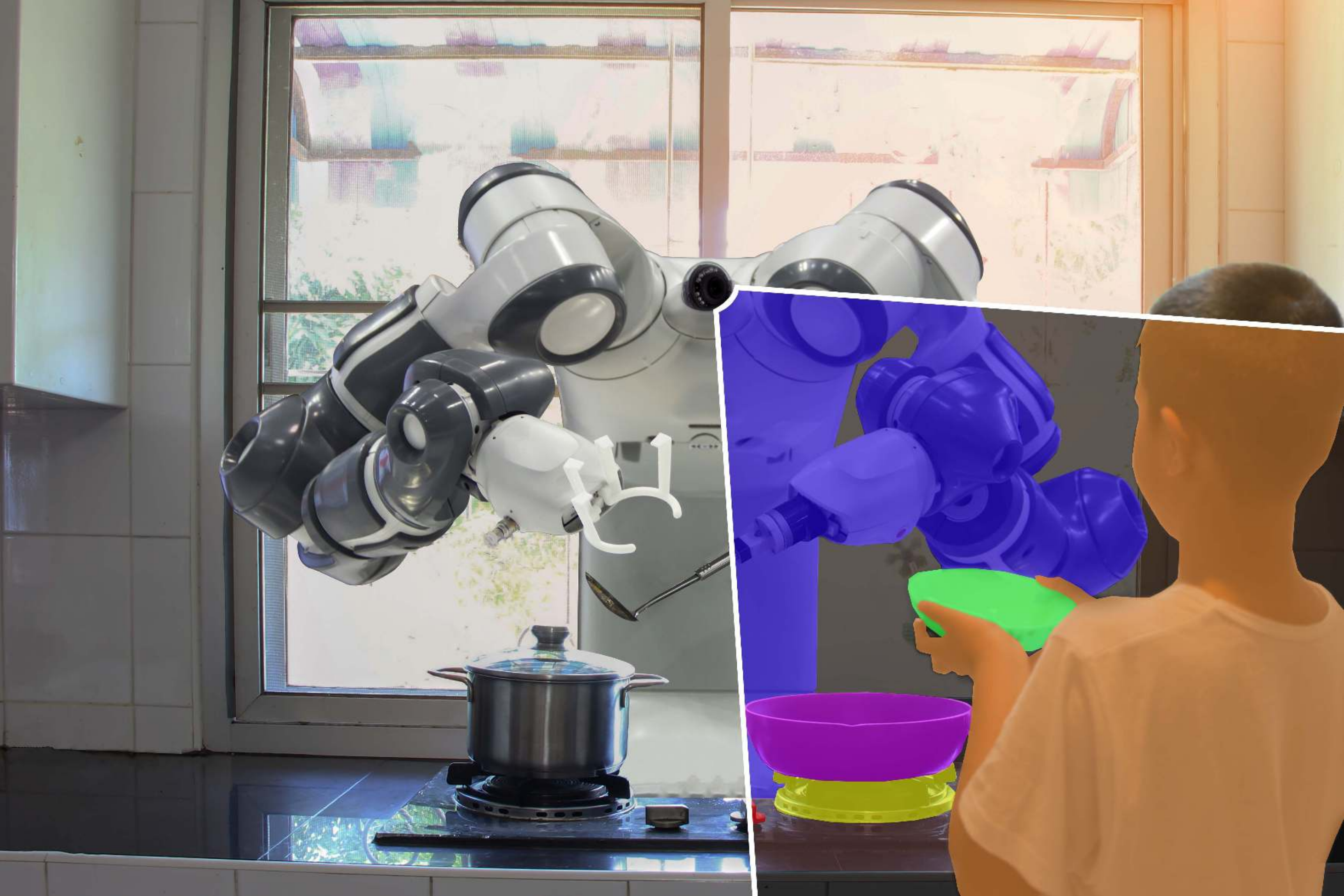} 
\end{tabular}
\vspace{-0.1cm}
\caption[Applications]{Autonomous cars, industry robots and home assistant robots are just some of the possible real world applications of UDA in semantic segmentation \footnotemark .}
\label{fig:applications}
\end{figure}

In Figure~\ref{fig:applications} we show three typical scenarios in which UDA for semantic segmentation could be highly valuable: namely, autonomous vehicles, industrial automation and domestic robots. 

\footnotetext{The images are modified version of pictures obtained with kind permission from Shutterstock, Inc. The original versions have been created (from left to right) by Scharfsinn, Monopoly919 and PaO\_Studio. }

\subsection{Outline}

In this paper, we mainly focus on analyzing and discussing deep UDA methods in semantic segmentation.
Recently, there has been a large number of studies related to this task. However, the motivating ideas behind these methods are different. To connect the existing work and hence to better understand the problem, we organize the current literature into some categories. 
We hope to provide a useful resource for the research of UDA in semantic segmentation. 

The rest of the survey is organized as follows:
in Section \ref{sec:problem} a concise and precise formulation of UDA for semantic segmentation is given outlining the various  stages at which the adaptation process may occur. Then, in Section \ref{sec:literature_review} we give an overview of the state of the art literature on the topic. We start from precursor techniques with weak supervision and then we propose a categorization based on the techniques employed to align the source and target distributions. In Section \ref{sec:datasets} we introduce a case study of synthetic to real unsupervised adaptation for semantic understanding of road scenes and we give an overview of the results of existing methods grouped by network architecture and evaluation scenario. In Section \ref{sec:conclusion} we conclude our review with some final considerations on the different adaptation techniques and we outline some possible future directions.

%% file: sections/problem_formulation.tex
\section{Unsupervised Domain Adaptation for Semantic Segmentation}
\label{sec:problem}

\subsection{Problem Formulation}
\label{sec:subsec:problem}
%

\begin{figure}[t]
\includegraphics[width=\textwidth]{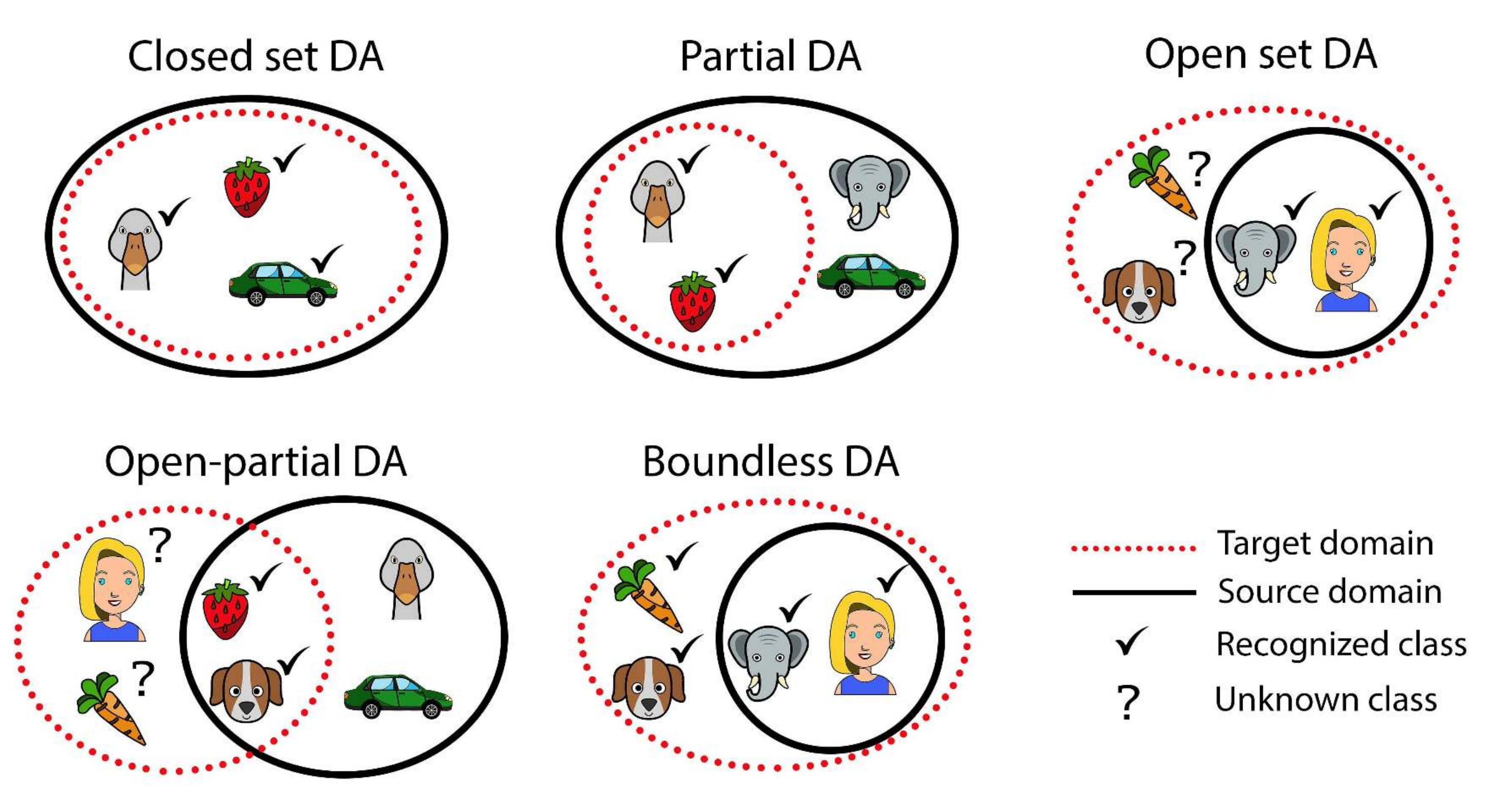}
\caption{Possible intersections between source and target domains.}
\label{fig:DA_sets}
\end{figure}

Image classification and image segmentation can both be reconducted to the problem of finding a function $h: \mathcal{X} \to \mathcal{Y}$ from the domain space $\mathcal{X}$ of input images to the label space $\mathcal{Y}$, that contains, respectively, the classification tags or the semantic maps.
From a mathematical point of view, it is possible to suppose that all real-world labeled images $(x, y) \in \mathcal{X} \times \mathcal{Y}$  are drawn from an underlying, fixed and unknown probability distribution $\mathcal{D}$  over $\mathcal{X} \times \mathcal{Y}$.
The search of the function $h$ should be limited to a predefined function space $\mathcal{H}$, called hypothesis class, chosen based on the prior knowledge on the problem.
In a supervised setting, a dataset of i.i.d. samples from $\mathcal{D}$ is used by the learner to find the best mapping $h \in \mathcal{H}$ (i.e., the solution that minimizes a cost function over the training set).  On the other hand, in DA, two different and related distributions over $\mathcal{X} \times \mathcal{Y}$, namely a source distribution $\mathcal{D}_S$ and a target distribution $\mathcal{D}_T$, are considered. A source domain training set $\mathcal{S}$ is sampled from $\mathcal{D}_S$ and a target domain training set $\mathcal{T}$ is sampled from $\mathcal{D}_T$ or from its marginal distribution over $\mathcal{X}$. 
The main purpose of DA is to use labeled i.i.d. samples from source domain $\mathcal{S}$ and labeled, or unlabeled, or a mixture of both, i.i.d. samples of the target domain $\mathcal{T}$ to find a hypothesis $h \in \mathcal{H}$ that performs well on the target domain $\mathcal{T}$. 
The DA task is supervised if labels in the target domain are available for all samples; it is semi-supervised if labels are available for just some samples; or it is unsupervised if the target samples are completely unlabeled (i.e., they are drawn from the marginal distribution of $\mathcal{D}_T$ over $\mathcal{X}$).  
Domain adaptation can be subdivided even further based on the categories (i.e., classes or labels) of the source ($C_S$) and target ($C_T$)  domains, and on the categories considered in the learning process ($C_L$): 
\begin{itemize}

\item \textbf{Closed Set DA}: all the possible categories appear in both the source and  target domains ($C_S = C_T$);
\item \textbf{Partial DA}: all the categories appear in the source domain, but just a subset appears in the target domain ($C_S \supset C_T$);
\item \textbf{Open Set DA}: some categories appear in the source domain and all categories appear in the target domain ($C_S \subset C_T$);
\item \textbf{Open-Partial DA}: some categories belong only to the source or to the target set and others belong to both sets ($C_S \neq C_T$ and $C_S \cap C_T \neq \emptyset$);
\item \textbf{Boundless DA}: an Open Set DA where all the target domain categories are learned individually ($C_S \subset C_T$ and $C_L = C_S \cup C_T$). 
\end{itemize} 
It is important to remark that in Open Set DA, usually, the categories of the target set that do not belong also to the source domain are learned by the model as an \textit{unknown} additional class, while in Boundless DA \cite{bucher2020buda} they are learned individually.
An overview of the aforementioned classification is given in Figure~\ref{fig:DA_sets}.

\input{sections/adaptation_spaces.tex}

%% file: sections/adaptation_spaces.tex
\subsection{UDA in Semantic Segmentation: Adaptation Spaces}

\begin{figure}[htbp]
\includegraphics[width=\textwidth]{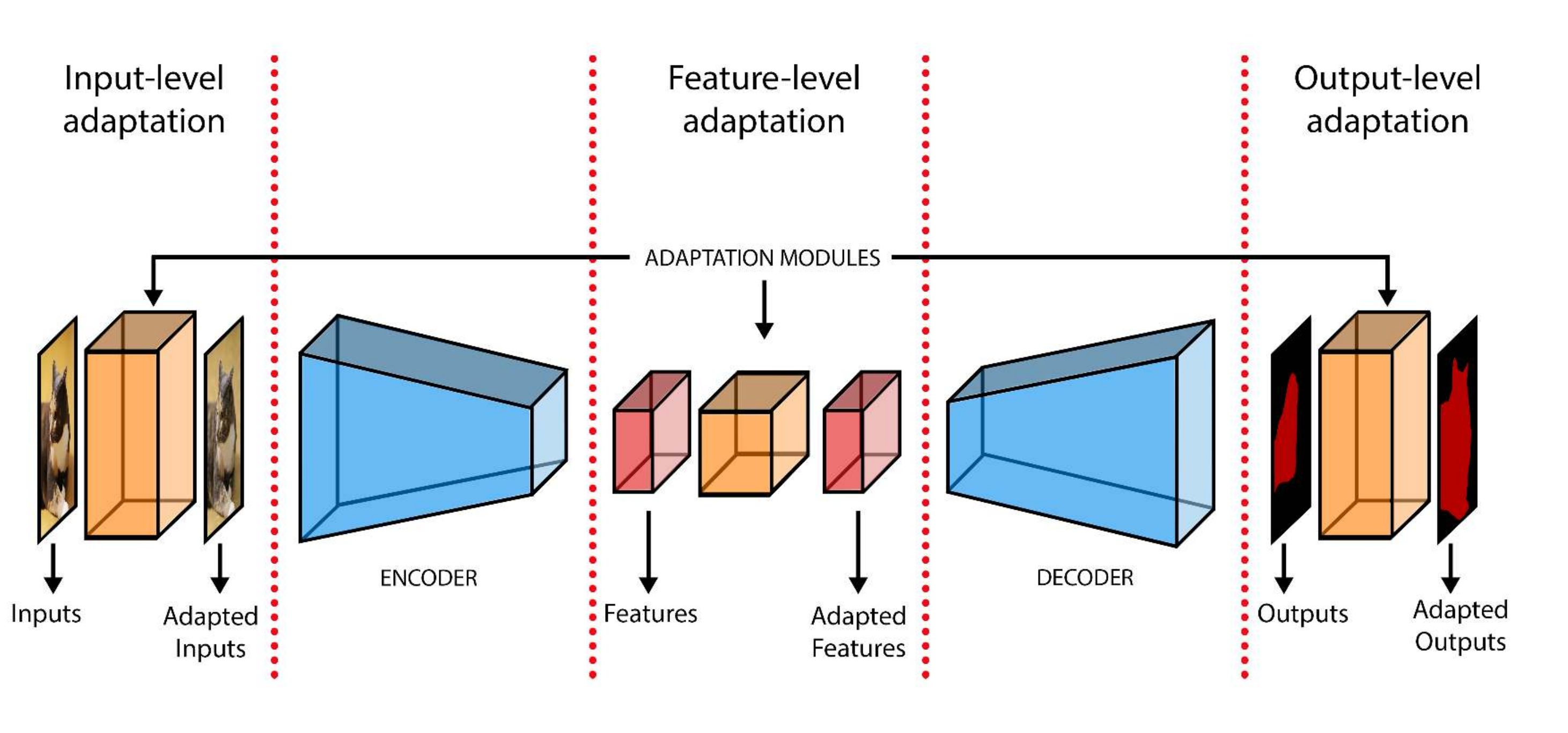}
\caption{Domain shift adaptation could be performed at different spaces: input-level, feature-level and output-level.}
\label{fig:adaptation_levels}
\end{figure}

As previously discussed, there exists a covariate shift phenomenon between source and target datasets, which prevents the network from yielding satisfactory results on the unsupervised target data.
Therefore, the primary strategy to tackle the domain adaptation problem is to bridge the gap existing between source and target distributions. 
In doing so, the performance drop that affects prediction models 
should lessen, thus allowing effective prediction whenever the original form of statistical discrepancy is successfully removed. 
In the following, a review of the different levels at which adaptation may be performed is presented, which will be also useful for the paper categorization in Section \ref{sec:literature_review}. A visual representation of the possible levels of adaptation is shown in Figure~\ref{fig:adaptation_levels}.
%
%
%

\textbf{Adaptation at the Input Level:} one way to proceed is to address the statistical matching at the input level to achieve cross-domain uniformity of visual appearance of the input image samples. Even if source and target images carry strong high-level semantic similarity in scene content and layout, inter-domain low-level statistical discrepancy, although mostly lacking semantic significance, is likely to result in an undesirable reduction of the prediction efficacy on target samples. 
In light of these considerations, a rich line of works have been focusing on style transfer techniques to close the marginal distributions of source and target images from original image-level sets. 
The common approach is to discover a function that maps source images to a new space, where the projected samples should carry an enhanced perceptual resemblance to the target ones. Then, the image segmentation network can access samples from the domain invariant input space during training. 
Recently, translation on the other way around has been explored as well, by which target images are transferred to the source domain before being fed to the segmentation network.\\
This strategy, despite being in principle completely task independent (it is usually performed in a stage detached from the training of the task predictor), 
is missing sufficient discriminative power 
when it is employed in its vanilla scheme, without any additional regularization constraints. 
Indeed, alignment of marginal distributions can be fully accomplished, and yet no semantic consistency may be preserved, with class-conditional distributions (not accessible at training time in the unsupervised target domain) still differing across domains. 
In other words, one might find many domain invariant representations, all lacking semantic discriminativeness to solve the segmentation task in the target domain. This for example could happen when objects of a certain class are mapped to different categories, which may be totally complying with the statistical 
alignment constraint, while, in fact, disregarding content preservation.
To bypass these issues, multiple approaches have been devised to enforce semantic consistency of image translations, for example by resorting to image reconstruction constraints, uniformity of segmentation predictions or ad-hoc engineered 
techniques to safely manipulate low-level statistics.

\textbf{Adaptation at the Feature Level:} a different approach is to seek for a distribution alignment of network latent embeddings. 
The core idea is to force the feature extractor to discover domain-invariant features, by adjusting the distribution of latent representations from source and target domains, both globally and class-wise. In this way, the network classifier should be able to learn to segment both source and target representations from the common latent space, by relying solely on the supervision from source data.
%
%
Compared to the classification task, in which feature domain adaptation has been successfully applied, semantic segmentation entails a much more complex and high-dimensional feature space, which should encode both local and global visual cues. 
Thus, alignment at feature level in its simplest fashion could be less effective in semantic segmentation, due to the structural and semantic complexity that feature embeddings possess, which is difficult to fully capture and 
handle (e.g., by an adversarial discriminator) \cite{yang2020fda}. 
In addition, even though adapted features should in principle retain semantic discriminativeness, they actually correspond to an intermediate representation in the segmentation process and 
there is no guarantee that the joint image-label distributions are aligned between domains, as unlabeled target images are drawn only from the marginal distribution. 
This can give rise to incorrect knowledge generalization to the unsupervised target representations. 
For such aforementioned reasons, feature adaptation has been adopted in semantic segmentation in combination with other complementary techniques or with specific arrangements to carefully overcome these major issues.

\textbf{Adaptation at the Output Level:} to avoid dealing with an excessively convoluted latent space, a different group of adaptation methods resort to the cross-domain distribution alignment over the segmentation output space.
While retaining enough complexity and richness of semantic cues, prediction maps from the segmentation network output (or the per-class outputs of the very last layers) identify a low-dimensional space where adaptation can be performed quite effectively, for example recurring to adversarial strategies. 
Moreover, label statistics over segmentation maps can be easily inferred over unlabeled target data, introducing a form of self-constructed weak supervision to the segmentation task. Source priors from label distribution can be profitably 
imposed in the adaptation process as well, as they usually involve high-level structural properties unbounded to the specific domain.

\textbf{Adaptation at Ad-Hoc Network Levels:} in addition to the aforementioned techniques, other works resort to a distribution alignment over ad-hoc spaces upon network activations. Such methods aim at better capturing high-level patterns essential to solve the segmentation task, and ultimately achieve an improved match of source and target embeddings, thanks to gradients flowing back through the segmentation network at different levels. Hence, the adaptation is not only restricted to a particular network level, i.e., at the end of the feature extraction network, but it is achieved at intermediate levels as well. 

%% file: sections/literature_review.tex
\section{Review of Unsupervised Domain Adaptation strategies}
\label{sec:literature_review}

\begin{figure}
\includegraphics[width=\textwidth]{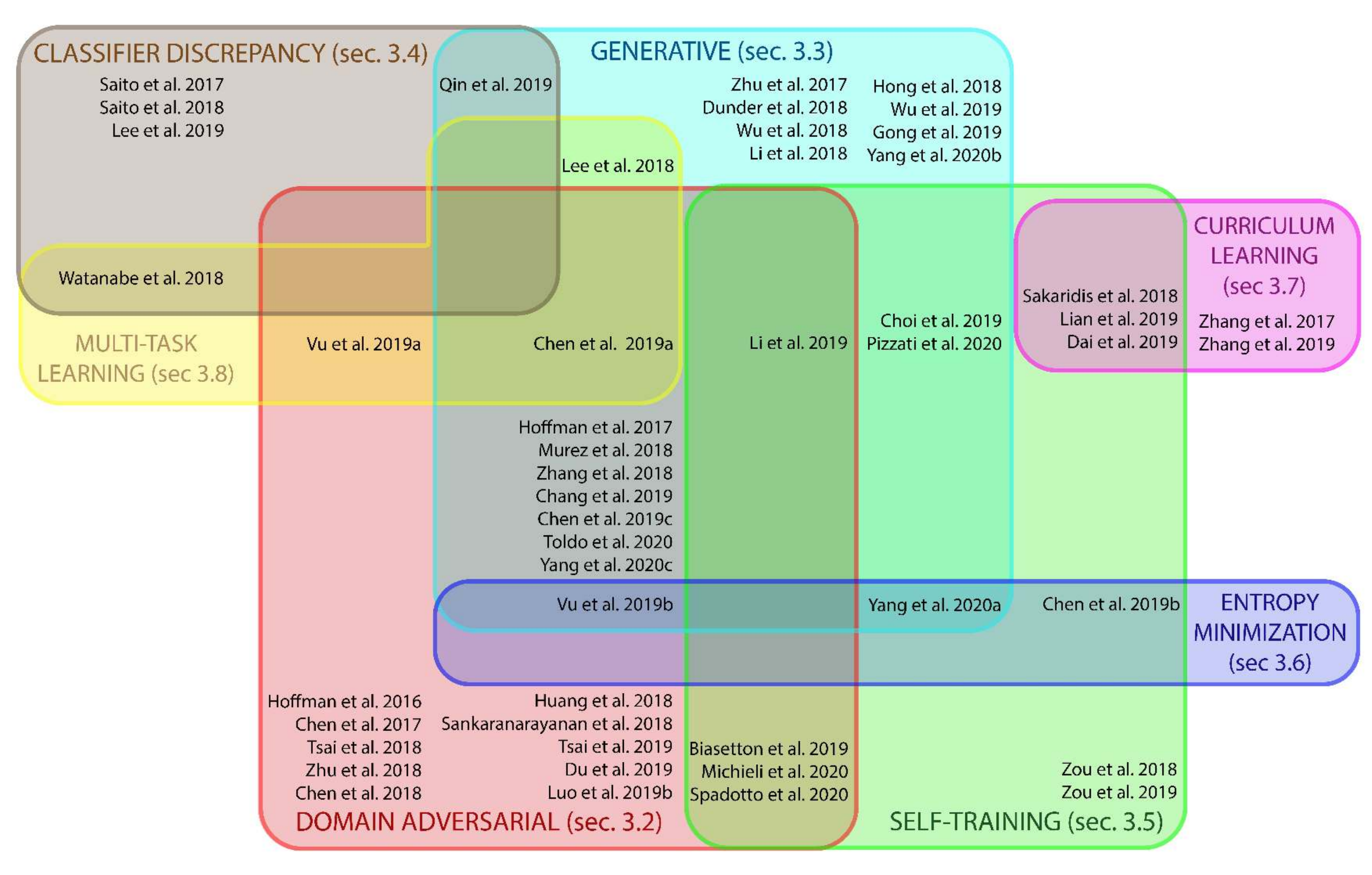}
\caption{Venn diagram of the most popular UDA strategies for semantic segmentation. Each method falls in the set representing the adaptation techniques used. Better viewed in color.}
\label{fig:papers_diagram}
\end{figure}

This  section reviews the most relevant approaches for Unsupervised Domain Adaptation in semantic segmentation. We start this section by presenting some weakly- and semi- supervised learning methods for semantic segmentation. Those are not strictly UDA approaches since they require some minimal supervision with annotations on typically simpler tasks, but have represented the starting point in dealing with the domain adaptation problem. Then, we grouped UDA approaches into $7$ main categories as shown by the visual overview in Figure~\ref{fig:papers_diagram}. 
Domain adversarial discriminative approaches (Section~\ref{sec:subsec:domain}) learn to produce data with a statistical distribution similar to the one of training samples via adversarial learning schemes. 
Generative-based approaches (Section~\ref{sec:subsec:generative}) typically use generative networks to translate  data between domains in order to produce a target-like training set from source data, or alternatively to translate the source data into a representation closer to target domain characteristics that can then be fed to the network.
Classifier discrepancy approaches in Section~\ref{sec:subsec:classifier} resort to multiple dense classifiers on top of a single encoder to capture less adapted target representations and, in turn, encourage an improved alignment of cross-domain features far from decision boundaries via an adversarial-like strategy.  
Self-training approaches in Section~\ref{sec:subsec:selftraining} propose to produce some form of pseudo-label (typically using some confidence estimation schemes to select the most reliable predictions) based on the current estimate to automatically guide the learning process (self-supervising it).
Entropy minimization methods in Section~\ref{sec:subsec:entropy} aim at minimizing the entropy of target output probability maps to mimic the over-confident behavior of source predictions, thus promoting well-clustered target feature representations. 
Curriculum learning approaches in Section~\ref{sec:subsec:curriculum} tackle one or more easy tasks first, in order to infer some necessary properties about the target domain (e.g., global label distributions) and then train the segmentation network such that the predictions in the target domain follow those inferred properties.
Multi-tasking methods in Section~\ref{sec:subsec:multitasking} solve multiple tasks simultaneously to improve the extraction of invariant features representation.
Finally, in Section~\ref{sec:subsec:newresearch} we conclude our digression with some considerations about recent interesting research directions to be further expanded in the future.

\subsection{Weakly- and Semi- Supervised Learning}

As earlier semantic segmentation works started from image classification techniques, also the domain adaptation task was originally tackled in the classification field, in that the first DA approaches for semantic segmentation have been developed by adapting DA methods for classification. However, approaches directly targeting the semantic segmentation task started to appear soon, taking into account the specific properties of the spatial components (completely missing in the classification methods) and of the dense (pixel-wise) task.
At the same time, unsupervised domain adaptation was historically preceded by techniques with weak or partial supervision, which are the focus of this section.

As we already mentioned, the training of a deep learning model for semantic segmentation requires a large amount of data with pixel-level semantic labels that are very expensive, difficult, frustrating and time-consuming to acquire. The problem is not so much relevant for other computer vision tasks like image classification and object detection because image-level tags or bounding boxes (that in this context are called \textit{weak} labels)  are much simpler to be obtained and large annotated collections are available.
This is the motivation behind many works that propose to use just weakly labeled samples to train a model in the segmentation task (weakly-supervised learning) or to use a mixture of many weakly labeled samples and few samples with the more expensive pixel-level semantic map (semi-supervised learning).

A first approach to solve the problem is to cast the weakly supervised semantic segmentation as Multiple Instance Learning problem as shown in \cite{vezhnevets2010towards} and \cite{pathak2014fully}. A Semantic Texton Forest (STF) has been used as base framework and an algorithm to estimate unobserved pixel label probabilities from image label probabilities has been introduced. Then, the structure of the STF has been improved through a new algorithm that uses a geometric context estimation task as regularizer in multi-task learning framework. \\ 
Another strategy, proposed in \cite{papandreou2015weakly}, is to implement an Expectation-Maximization (EM) method to train a deep network for the semantic segmentation task in a weakly- and semi-supervised setup. The algorithm alternates between estimating the pixel-level annotations (constrained on the weak annotations) and the optimization of the segmentation network itself. 
In \cite{pathak2015constrained} Constrained CNNs (CCNNs) have been introduced as framework to incorporate weak supervision into the training. Linear constrains are added in the output space to describe the existence and expected distribution of labels from image level tags and a new loss function is introduced to optimize the set of constrains. \\
In \cite{wei2016stc} a simple to complex framework has been introduced for weakly-supervised semantic segmentation. In the paper a distinction is made between simple and complex images: the former include a single object of just one category in the foreground and a clean background, the latter can have multiple objects of multiple categories with a cluttered background. First, salient object detection techniques are used to compute semantic maps from weak-annotated simple images and, then, starting from these, three different networks are trained, sequentially, in order to gradually enable the segmentation of complex images.

In \cite{hong2015decoupled} a semi-supervised approach is presented, whose architecture is composed of three main structures: a classification network, a segmentation network and some bridging layers that interconnect the two networks. The proposed training is decoupled: first, the classification network is trained with weakly-annotated samples, then, the bridging layers and the segmentation network are jointly trained with the strong-annotated samples. The input image is first fed to the classification network, then the bridging layers extract from an intermediate layer of the classification network a class-specific activation map that is used as input for the segmentation network. In this way, it has been possible to reduce the number of parameters of the segmentation network and to make a training with just few semantic-annotated samples possible. In fact, relevant labels and spatial information are captured from the classification network and refined by the bridging layers and the task of the segmentation network is widely simplified. \\
In \cite{dai2015boxsup} an iterative procedure has been proposed to train a segmentation network just with bounding-boxes annotated samples. First, region proposal methods are used to generate many candidate segmentation masks (that are fixed throughout the training) for each image. An overlapping objective function is defined to pick the candidate mask that overlaps the ground truth bounding box as much as possible with the correct label. At every iterative step,  one candidate mask is selected for each bounding box and then the resulting semantic labels are used to train the segmentation network. The outputs of the segmentation network are then used to improve the choice of the candidate labels for the next step through a feedback channel. After every iteration, the selected candidate labels and the segmentation network outputs both improve together. \\
Generative adversarial networks have  proven to be effective in this filed starting from \cite{souly2017semi}, where the discriminator network is modified to accomplish the task of semantic segmentation. The discriminator assigns to every pixel of the input image either a label of one of the semantic classes or a fake label. The discriminator is trained with fake (generated) data, unlabeled data for regularization purposes, and with labeled data with pixel-level semantic maps. Another proposed solution is to employ conditional GANs and incorporate weak image-level annotation both at the generator and at the discriminator inputs in a weakly-supervised setup.\\
Many approaches of self-supervised learning have been proposed starting from \cite{kolesnikov2016seed}. The common rationale is the exploitation of inferred pixel-level activations as pseudo ground-truth in order to obtain more accurate pixel-level segmentation maps. In \cite{huang2018weakly} an image classification network with classification activation maps has been used. The authors highlight how the discriminative regions, using that method, are small and sparse and they propose to use them as seed cues. Then, the regions are expanded to neighboring pixels with similar features (for example color, texture or deep features) with a classical Seeded Region Growing (SRG) algorithm to obtain accurate pixel-level labels that are used to train a segmentation network. The output of the segmentation network is used by the SRG algorithm to compute the similarity between the seed and the adjacent pixels. So, at every iteration, the segmentation network and the dynamic labels computed with SRG improve together.
A similar approach that introduces a new adversarial erasing method for localizing and expanding object regions progressively is presented in \cite{wei2017object}. Other self-learning based techniques are presented in \cite{ahn2018learning}, \cite{lee2019ficklenet} and \cite{ahn2019weakly}. \\
A more general framework to transfer knowledge across tasks and domains is presented in \cite{ramirez2019learning}. 
Assuming to have two tasks 
and two domains
, the proposed method works in 4 steps: (1) a single task network is trained on samples of both domains to solve the first task, in order to find a common feature representation for the domains, (2) a second network is trained  to solve the second task just on the first domain, (3) a third network is trained on the first domain to map deep features suitable for the first task into features to be used for the second task (4) finally, the third network is used to solve the second task on the second domain. 
This framework enables to adapt from a synthetic domain to a real one for the image segmentation task using depth maps of both domains. 
The depth maps can be considered weak annotations with respect to semantic maps because their acquisition is easier thanks to depth cameras and 3D scanners. 

\input{sections/adaptation_techniques.tex}

\input{sections/adaptation_newresearchdirections.tex}

%% file: sections/adaptation_techniques.tex
\input{sections/domain_adversarial_discriminative.tex}

\input{sections/generative_based_approaches.tex}

\input{sections/classifier_discrepancy.tex}

\input{sections/self_training.tex}

\input{sections/entropy_minimization.tex}

\subsection{Curriculum Learning}
\label{sec:subsec:curriculum}
Another research area regards curriculum learning approaches, where some easy tasks are solved first, inferring some important and useful properties related to the target domain. Then, this information is used to support the training of a network dealing with a more challenging task, like image segmentation. This family of approaches shares many similarities in spirit with self-training. The main difference between the two approaches lies in the content of the pseudo-labels. While in the self-training approaches the pseudo-label is an estimate of the desired annotation on the target set and it is used as such during training, in curriculum approaches the pseudo-label is represented by some inferred statistical properties of the target domain (different from the labels for the task) and the network is trained to reproduce such inferred properties in the target predictions.

The first work of this family is \cite{zhang2017curriculum} and its extension \cite{zhang2019curriculum}, where a couple of easy tasks that are less sensitive to domain discrepancy are solved: namely, the label distribution over global images and the label distribution over local landmark superpixels. The former property is evaluated in the source domain, as the number of pixels in the labels associated to each category, normalized by the total number of pixels. On the other hand, target labels are not available in unsupervised domain adaptation and consequently a machine learning model should be trained on the source domain to estimate them. 
In the papers, it is argued that this task can be solved more easily than image segmentation and that the results can be used to guide the adaptation of the segmentation task. To estimate the first property on the target domain a logistic regression model is employed. 
While the first property is useful to guarantee that the ratio among different categories matches the ones of the target domain, samples with semantic maps not following the estimated label distribution on the target domain are still penalized. To solve this problem, a second clue is introduced. Images are divided into superpixels and an SVM classifier is used to select the most representative anchor superpixels and the label distribution is estimated over them. 
The final objective is a mixture of the pixel-wise cross-entropy of the source samples and the cross-entropy on the two properties on the target domain discussed before. 
In \cite{sakaridis2018model} and \cite{dai2019curriculum} a technique to adapt the domain of a segmentation model from clear weather to dense fog images is introduced. A novel method, called Curriculum Model Adaptation (CMAda), is proposed to gradually adapt the model to segment images with incrementally growing amount of fog.  A new method to add synthetic fog to images and a new fog density estimator are introduced. 
It is important to remark that the fog generator has a tunable parameter $\beta$ that controls the density of the fog to add to the images. This made possible to generate samples from the dataset Cityscapes with different synthetic fog density and to use them to train an AlexNet model \cite{krizhevsky2012imagenet} to perform a regression problem to discover $\beta$ from images. The trained fog density estimator, then, can also be used to estimate the fog density of real foggy images. The algorithm presented starts from a source domain of clear weather images and progress through intermediate target domains of incrementally denser fog, and, finally, reaches the target domain of dense fog images. During training, the labels of source domain and of synthetic fog images are available, while the real foggy images are unlabeled. The segmentation model, initially, is pretrained with supervision on the source domain and then, with as many adaptation training steps as the number of denser fog steps, it is gradually shifted towards the target domain. 
Starting from the assumption that images with lighter fog are easier to segment, the model of current step is used to evaluate the labels for real foggy images with intensity less than the $beta$ of the current step. Then these samples are used together with images with synthetic fog of density $beta$ of the next step to train the model with supervision. Iterating this process for all the steps towards the target dataset, the model adapts, in an unsupervised way (labels of the real foggy images are not used), to segment real dense fog images. 
In \cite{lian2019constructing} the connection between curriculum learning and self-training is highlighted and a method (called self-motivated pyramid curriculum domain adaptation, PyCDA) that uses and merges both techniques is presented. 
The authors remind that in self-training there are two main training steps that alternates: (1) the evaluation of pseudo-labels for the target domain and (2) the supervised training of the segmentation network with the labeled source domain images and with the target domain images with pseudo-labels. In curriculum learning there are also two steps that alternates: (1) the inferring of properties of the target domain (e.g., frequency label distributions over global images or image regions, like superpixels) and (2) the update of the network parameters using the labeled source domain and the target domain inferred properties. In PyCDA the two approaches are merged: the pseudo-labels used in self-training are considered as a property of the curriculum approach. The papers also substitute the superpixels used in \cite{zhang2017curriculum} with small squared regions to improve the algorithm efficiency and also all the curriculum properties are inferred with the segmentation networks itself and additional models (for example SVMs or logistic regression models) are not needed. 

\subsection{Multi-Tasking}
\label{sec:subsec:multitasking}

Some works exploit additional types of information available in the source domain dataset, e.g., depth maps, to improve the performance in the target domain. In other words, the models are trained to solve additional tasks (for example depth regression) simultaneously to image segmentation in order to build an invariant and generic embedding of the images.
In \cite{lee2018spigan} the authors highlight that when the source domain is made of synthetic data 
we could include other information about the dataset samples beyond the semantic maps, e.g. depth maps. This is called Privileged Information (PI) and it includes all properties that may be useful for the training. 
The method proposed in \cite{lee2018spigan} is called Simulator Privileged Information and Generative Adversarial Networks (SPIGAN), which uses an adversarial learning scheme performing source-to-target image translation  together with a  network trained on source images and on adapted images that tries to predict their privileged information (e.g., the depth map). In particular the PI is used as regularization for the domain adaptation.
A different use of the extra depth information in the source domain to enhance the appearance features and improve the alignment of the source-target domains is presented in \cite{vu2019dada}. The method introduced is called Depth-Aware Domain Adaptation (DADA) and includes a specific architecture and a learning strategy. The architecture starts from an existing segmentation network and includes some extra modules to predict the monocular depth and to feed the information of this task back in the main stream. Residual auxiliary blocks are used for this purpose. To perform domain adaptation, images from source and target domains are fed to the network to compute the class-probability and depth maps. Then, the former is processed into self-information maps and merged to the latter 
to produce depth aware maps. Finally, these maps are used in an adversarial training to adapt the source domain. It is important to remark that the depth information is not used as regularization, but it is directly considered while deriving prediction for the main task. In the paper is argued that this is a more explicit and more useful way to exploit the depth information than the method presented in \cite{lee2018spigan}.

A third different use of the depth maps is introduced in \cite{chen2019learning} where a method called Geometrically Guided Input-Output Adaptation (GIO-Ada) is presented. The geometric information is exploited to improve the adaptation both at the input-level and at the output-level. The former adaptation tries to reduce the visual differences of the images of the source and target domains. A transform network accepts as input source images together with their semantic maps and depth maps to compute adapted images, visually similar to images of the target domain. A discriminator is used in an adversarial learning with the transform network to distinguish real target domain images from adapted ones. The main contribution of the paper in this adaptation is the use of semantic maps and depth maps as additional inputs for the transform network. The output-level adaptation is built with a task network that computes, for each input, the semantic map and the depth map. Such outputs are fed to an additional discriminator which tries to distinguish whether they were computed from a real or adapted image. 
In \cite{watanabe2018multichannel} a network composed of a feature generator followed by two classifiers (that computes semantic maps) has been adopted and a maximum classifier discrepancy approach is used for the unsupervised adaptation from a synthetic source domain to a real target domain. Two techniques are presented to improve the performance of the network: a data-fusion approach and a multi-task one. The former merges the RGB image information and the depth information and use the result as input of the network. 
In the latter only RGB images are used as inputs, however 3 tasks are solved simultaneously by different networks after the feature generator to boost the overall performance of the network in the target domain: namely, semantic segmentation, depth regression and boundary detection.

%% file: sections/domain_adversarial_discriminative.tex
\subsection{Domain Adversarial Discriminative}
\label{sec:subsec:domain}

\textbf{Adversarial Learning:} Adversarial learning has been introduced in the form of Generative Adversarial Networks (GANs) \cite{goodfellow2014generative} to address a generative objective (i.e., generating \textit{fake} images similar to real world ones). 
Solving the generative task can be thought as seeking for 
the evaluation of the unknown probability distribution from which the training data has been generated. 
In the generative context, the introduction of adversarial learning has been ground-breaking, as explicit modeling of the underlying target distribution is not required and, more importantly, no specific 
objective is needed to train the model. 
In the adversarial scheme, a generator has to learn to produce data with the same statistical distribution of training samples. To do so, it is paired with a discriminator, which has the goal of understanding whether input data comes from the original set or, instead, it has been generated. At the same time, the generator is optimized to fool the discriminator by producing samples that resemble the original ones. 
In the end, the statistics of generated data should match the one of the training set.
The GAN model is capable of learning a structured loss in the form of a learnable discriminator, which guides the generative network in its optimization procedure. For this reason, the objective function can be thought as automatically adapting to the specific context, removing, in fact, the necessity to manually design complex losses. Therefore, the adversarial learning scheme introduced in \cite{goodfellow2014generative} can be extended under careful adjustments to address multiple tasks that would normally require different types of application-specific objectives. 

\textbf{Feature Adversarial Adaptation:} Aiming at exploiting the statistical matching that can be achieved by the GAN model, adversarial learning has been successfully extended to the domain adaptation task \cite{ganin2015unsupervised,ganin2016domain,tzeng2017adversarial}. 
In particular, the real-fake discriminative network in the original adversarial framework is revisited, turning into a source-target domain classifier. Thus, while the segmentation network is trained with source supervision to achieve discriminativeness over the semantic segmentation task, the supervisory signal provided by the domain discriminator should guide the predictor to reach domain invariance and reduce the otherwise intrinsic bias towards the source domain. 
In other words, a measure of domain discrepancy is simultaneously learnt and minimized within the adversarial competition.
\\
Although the adversarial adaptation strategy has been originally introduced for the image classification task \cite{ganin2015unsupervised,ganin2016domain}, it has been later extended to image semantic segmentation. Hoffman et al. \cite{hoffman2016fcns} have been the first to address domain adaptation in semantic segmentation, and  they  resort to an adversarial approach. In particular, they devise a global domain adversarial alignment, based on a domain discriminator taking as input the feature representations from intermediate activations of the fully convolutional segmentation network. 
%
%
In addition, 
they propose a category specific distribution alignment, which is accomplished by enforcing image-level label distribution constraints on target predictions inferred from source annotations, under the assumption that high-level scene layout is in general shared among source and target images.
\\
Following a similar approach to \cite{hoffman2016fcns}, many works have further resorted to adversarial alignment of network latent embeddings \cite{chen2018road, zhang2018fully, li2019bidirectional, huang2018domain, hoffman2017cycada, chen2019crdoco, luo2019significance, toldo2020unsupervised, chen2017no, du2019ssf}.
As previously discussed, the domain discriminator is able to infer a structured loss to capture global distribution mismatch of cross-domain image representations. However, 
global alignment of marginal distributions does not necessarily result in class-wise correct semantic knowledge transfer from source to target representations. Thus,  adversarial learning is commonly employed in  more complex frameworks working also on the internal feature representations of the network, comprising multiple complementary modules to achieve a more effective adaptation. 
For example, Chen et al. \cite{chen2018road} use an additional target guided distillation loss by matching network activations from target inputs during the training phase with those from a pre-trained version on the  ImageNet dataset \cite{Deng2009Imagenet}. In this way, they argue that overfitting to source data is decreased. Moreover, the feature adversarial adaptation is enforced independently over different spatial regions of the input image, thus exploiting the underlying spatial structure of input scenes.    
Zhang et al. \cite{zhang2018fully}, instead, boost the feature-level adaptation performance by providing the domain discriminator with an Atrous Spatial Pyramid Pooling (ASPP) module \cite{chen2018deeplab} to capture multi-scale representations. 
More recently, Luo et al. \cite{luo2019significance} propose a significance-aware information bottleneck (SIB) to filter out task-independent information encoded inside feature representations, so that, when enforcing adversarial adaptation, only domain invariant discriminative cues are preserved. They also introduce a significance-aware module to help the prediction of less frequent classes, which may be penalized by the information bottleneck in its original form. 
\\
Another group of researches \cite{hoffman2017cycada,chen2019crdoco,li2019bidirectional,toldo2020unsupervised} combines a generative approach (which will be extensively discussed in the following section \ref{sec:subsec:generative}), with the adversarial feature alignment. In particular, source and target marginal distributions are matched in the input image space by a source-to-target image-to-image translation function, and then cross-domain latent representations are further brought closer by matching source original and target-like source embeddings in a domain adversarial fashion. 
\\
To accomplish category-wise adaptation, some works \cite{chen2017no, du2019ssf} revisit the original approach of Hoffman et al. \cite{hoffman2016fcns} by assisting the global distribution alignment with class-wise adversarial learning. Chen et al. \cite{chen2017no} propose to exploit multiple feature discriminators (one for each class), so that negative transfer among different classes in the domain bridging process should be effectively avoided. In addition, due to lack of ground-truth maps, they use grid-level soft pseudo labels from network predictions to compute the target adversarial loss. 
Recently, Du et al. \cite{du2019ssf} propose a similar class-wise adversarial technique, which is improved by imposing independence during the optimization of the multiple discriminators. They argue that soft labels lead to incorrect adaptation on class boundaries, where different class discriminators may provide their guidance simultaneously. Finally, they devise an additional module to adaptively re-weight the contribution of each class component in the adversarial loss, in order to avoid the inherent dominance of classes with higher prediction probability, which turns out to be more easily well-adapted across the domains.

Different from the aforementioned techniques, other works \cite{zhu2018penalizing, murez2018image, sankaranarayanan2018learning} seek for domain alignment inside the feature space by applying a reconstruction constraint to ensure that latent embeddings possess enough information to recover the input images from which they have been extracted. 
To this end, adversarial learning is applied on the reconstruction image-level space. To achieve cross-domain feature distribution alignment, the feature extractor is trained to yield latent representations that can be projected back to both source and target image spaces indistinctly. 
In these frameworks the backbone encoder of the segmentation network plays a min-max game against the domain discriminator. The encoder, indeed, tries to fool the discriminator on the actual originating feature's domain, 
by looking at the corresponding reconstructed images projected back into the image space. 
In other words, the objective is to learn source (target) features that can successfully generate target-like (source-like) images to promote domain invariance of those representations.

\textbf{Output Adversarial Adaptation:} To avoid the complexity of high-dimensional feature space adaptation, a different line of works \cite{tsai2018learning, chen2018geometric, chang2019all, luo2019taking, yang2020label, biasetton2019unsupervised, michieli2020adversarial, spadotto2020unsupervised} resort to adversarial adaptation on the low-dimensional output space spanned by the segmentation network, which is still expected to encode enough semantic information to allow effective adaptation. 
A domain discriminator is provided with prediction maps from source and target inputs and it is optimized to discern the domain they originate from. Conversely, the segmentation network has to fool it by aligning the distribution of predicted dense labels across domains.
Tsai et al. \cite{tsai2018learning} are the first to propose this type of adaptation: in order to improve the signal flow from the adversarial competition through the segmentation network, they deploy multiple dense classification modules at different depths upon which as many output-level discriminators are applied. 
\\
Following the technique proposed in \cite{tsai2018learning}, other works adopt the output space adversarial adaptation in combination with additional modules. For example, 
Chen et al. \cite{chen2018geometric} combine semantic segmentation and depth estimation to boost the adaptation performance. In particular, they provide the domain discriminator with segmentation and depth prediction maps jointly, in order to fully exploit the strong correlation between the two visual tasks. 
Moreover, Luo et al. \cite{luo2019taking} enhances the adversarial scheme by a co-training strategy that highlights regions of the input image with high prediction confidence. In this way, the adversarial loss can be effectively tuned by balancing the contribution of each spatial unit, so that more focus is directed towards less adapted areas.
\\
Other works \cite{biasetton2019unsupervised, michieli2020adversarial, spadotto2020unsupervised} revisit the adversarial output-level approach. In particular, they utilize a discriminator network which has to distinguish between source ground-truth maps and generated semantic predictions from either source and target data. In doing so, the cross-domain statistical alignment is not directly performed, but forcing the segmentation network output to be distributed as ground-truth labels for both source and target inputs leads to an indirect yet effective alignment between the two domains.

Recently, new approaches \cite{vu2019advent,vu2019dada,tsai2019domain} have been proposed based on the extraction of meaningful patterns from the segmentation output space to be exploited in the adaptation process. 
This is done to explicitly 
guide the domain discriminator towards a more functional and significant insight of source and target representations, and thus to ultimately achieve a better alignment. 
\\
On this regard, Vu et al. \cite{vu2019advent,vu2019dada} devise an entropy-minimization strategy (which will be described more in detail in section \ref{sec:subsec:entropy}) to promote more confident target predictions. They propose an indirect approach relying on the adversarial alignment of the statistics of self-information maps computed on top of source and target predictions. In particular, a domain discriminator has to detect whether a weighted self-information map comes from a source or a target prediction, whereas the segmentation network, trying to deceive the discriminator, is forced to produce low-entropy target maps as to mimic source confident ones. This process effectively pushes decision boundaries away from high-density regions in the representation space.
\\
With a different approach, Tsai et al. \cite{tsai2019domain} construct a clustered space over the output prediction space by adding a patch clustering module that discovers patch-wise modes on segmentation maps. First, the module is supervisedly trained on source data by leveraging the available annotations, then it is exploited to achieve a patch-wise distribution alignment by enforcing adversarial cross-domain adaptation between its clustered source and target representations.
The idea behind this approach is to capture high-level structured patterns, that are essential to solve the semantic segmentation task, to be provided to the domain discriminator for an improved domain statistical alignment. 
Thus, the achieved domain uniformity on a patch-level should ensure, in principle, that the segmentation task can be effectively solved also in the target domain.

%% file: sections/generative_based_approaches.tex
\subsection{Generative-Based Approaches}
\label{sec:subsec:generative}

Unsupervised image-to-image translation is a class of generative techniques where the objective is to learn a function that maps images across domains, relying solely on the supervision provided by unpaired training data sampled from the considered domains. The idea is to extract characteristics peculiar to a specific set of images and transfer those properties to a different data collection. 
In a more formal definition, the image-to-image translation task aims at discovering a joint distribution of images from different domains.
Notice that, since the problem is, in fact, ill-posed, as an infinite set of joint distributions can be inferred
from the marginal ones, appropriate constraints must be applied to obtain acceptable solutions.
%
%

Image-to-image translation can be effectively exploited in domain adaptation:  discovering the conditional distribution of the target set with respect to the source one, should allow, in principle, to bridge the statistical gap between source and target pixel-level statistics, thus removing the original covariate shift responsible of the classifier performance drop.
The goal, in fact, is to transfer visual attributes from the target domain to the source one, while preserving source semantic information.
Following this idea, many works have proposed an input-level adaptation strategy based on a generative module that translates images between source and target domains. 
Despite the wide range of different approaches, all these works share the same idea of achieving a form of domain invariance in terms of visual appearance, by mitigating the cross-domain discrepancy in image layout and structure. 
This allows to learn a segmentation network on translated source domain data (that should have a target-like statistical distribution) allowing to make use of source annotations.

A considerable amount of research \cite{hoffman2017cycada,chen2019crdoco,toldo2020unsupervised,zhou2020uncertainty,li2019bidirectional,murez2018image,qin2019generatively,li2018semantic,yang2020phase,gong2019dlow} has been resorting to the successful CycleGAN \cite{zhu2017unpaired} unsupervised image-to-image translation framework to accomplish input-level domain adaptation. 
%
%
The framework proposed by Zhu et al.\ \cite{zhu2017unpaired} is built on top of a pair of generative adversarial models, concurrently performing conditional image translation between a couple of domain sets, in both the source-to-target and target-to-source directions. The two adversarial modules are further tied by a cycle-consistency constraint, which encourages the cross-domain projections to be one the inverse of the other. This reconstruction requirement is essential to preserve structural geometrical properties of the input scene, but provides no guarantees about the semantic consistency of translations. In fact, while retaining geometrical coherence, the mapping functions could completely disrupt the semantic classification of input data.

Taking this into account, a number of works \cite{hoffman2017cycada,chen2019crdoco,toldo2020unsupervised,zhou2020uncertainty,li2019bidirectional} address semantic consistency by taking advantage of the semantic discriminative capability of the segmentation network.
In particular, cross-domain image translations are forced to preserve semantic content as perceived by the semantic predictor, which represents a measure of semantic discrepancy between an original image and its translated counterpart, that is minimized in the optimization of the translation network. 
Still, being the prediction maps intrinsically flawed, especially in the target domain where annotations are missing, the inaccurate semantic information provided to the generative module could hurt the learning of the image projections. Thus, some works propose to simultaneously optimize the generative and discriminative framework components in a single stage \cite{toldo2020unsupervised}, or even split the segmentation network into separate source and target predictors \cite{chen2019crdoco}.
%
Li et al.\ \cite{li2019bidirectional} further extend the CycleGAN-based adaptation strategy formulating a bidirectional learning framework. The image-to-image translation and segmentation modules are alternately trained, in an optimization scheme by which each module is provided with positive feedback from the other. The segmentation network benefits from the target-like translated source images with original supervision, while the generative network is aided by the predictor in retaining semantic consistency. This closed-loop structure effectively allows for a progressive adaptation, with both image-to-image translations quality and semantic prediction accuracy gradually enhanced.
\\ 
Other works \cite{li2018semantic,yang2020phase} resort to different approaches to provide semantic-awareness to the CycleGAN-based adaptation. 
Li et al.\ \cite{li2018semantic} propose to assist the cycle-consistent image-to-image translation framework by a soft gradient-sensitive loss to preserve semantic content in the cross-domain projection focusing on semantic boundaries. 
The idea behind this approach is that, no matter how low-level visual attributes change between domains, the edges defining semantic uniform regions should be easily detectable, regardless the distribution the image is drawn from. 
Thus, a gradient-based edge detector should discover consistent edge maps between original images and their transformed versions. 
In addition, following the intuition that semantically different regions of an image should face a different adaptation, they devise a semantic-aware discriminator structure. 
In doing so, the discriminator can semantically-wise evaluate resemblance between original and translated samples.
\\
Very recently, Yang et al.\ \cite{yang2020phase} introduce a phase consistency constraint to the CycleGAN pixel-level adaptation module, observing that the semantic content of an image is mostly encoded in the phase of its Fourier transform, whereas alterations of the amplitude to the representation in frequency does not change its composition.   
\\
With a different adaptation perspective, Gong et al.\ \cite{gong2019dlow} adapt the CycleGAN model to generate a continuous flow of domains ranging from source to target ones, by conditioning the generative networks with a continuous variable representing the domain.   
The reason behind the retrieval of intermediate domains spanning between the two original ones is to ease the adaptation task, by progressively characterizing the domain shift affecting the input data distributions. Moreover, they suggest that resorting to target-like training data from diverse target-like domain distributions improves the generalization capability of the segmentation network.

To reduce the computational burden of the bi-directional structure of CycleGAN (which entails a total of at least four neural networks to be added to the semantic predictor) other works \cite{chen2018geometric,lee2018spigan,choi2019self,hong2018conditional} discard the backward source-to-target projection branch, seeking for a more light-weight input-level adaptation module, still based on generative adversarial framework. 
The translation consistency is granted, for example, by the correlation to a related task (e.g. depth estimation) \cite{chen2018geometric,lee2018spigan}, which is jointly addressed with the semantic segmentation. 
Choi et al.\ \cite{choi2019self}, instead, improve the generator of the original GAN framework with feature normalization modules at multiple depths to provide style information to source representations, whereas source content is preserved. Furthermore, a semantic consistency loss from a pre-trained segmentation network promotes coherence of image translations, 
providing, in fact, a regularizing effect in absence of the cycle-consistency one.
Hong et al.\ \cite{hong2018conditional} use a conditional generative function to model the residual representation between source and target feature maps, which is optimized in an adversarial framework.
In doing so, they avoid any reliance on a shared domain-invariant latent space assumption, which may be not satisfied due to the highly structured nature of semantic segmentation. 
The generator takes as input low-level source feature maps, together with a noise sample, and is encouraged to produce high-level feature maps with target-like distribution by a discriminator, that expresses a measure of statistical distance between original and reproduced target representations.
Both source original and domain-transformed representations are provided to a dense classifier to compute the cross-entropy loss.

In order to lessen the bias towards the  source domain, Yang et al.\ \cite{yang2020label} resort to the target-to-source image-to-image translation, in place of the more common source-to-target one, generally employed to generate a form or target supervision from source translated data.
The source-like target images are then employed in the supervised training of the predictor thanks to pseudo-labeling. In addition, training the segmentation network directly in the source domain allows to fully exploit the original source annotations, avoiding the risk of semantic alterations which may happen in the source-to-target pixel-level adaptation scenario.
Moreover, to align feature representations between domains, they introduce a label-driven reconstruction network. However, differently from the feature-based reconstruction techniques \cite{zhu2018penalizing, murez2018image, sankaranarayanan2018learning} (section \ref{sec:subsec:domain}), the generative recreation of input images is performed starting from semantic maps from the segmentation output. In doing so, they seek to guide a category-wise alignment of the segmentation network embeddings, since reconstructions that deviate from their target are penalized, thus providing semantic consistency to network predictions.

A different category of adaptation strategies explores style-transfer techniques to achieve image-level appearance invariance between source and target domains. 
These approaches are based on the principle that every image can be disentangled into two separate representations, namely content and style. 
As the style encodes low-level domain-specific texture information, the content expresses domain-invariant high-level structural properties. 
Thus, being able to combine style properties from target data with semantically preserving source content should effectively allow for the construction of target-distributed training data, still retaining original source annotations. 
Some techniques \cite{chang2019all,pizzati2020domain} involve content and style decomposition in the latent space. Translating a source image, then, means extracting its feature content representation and recombine it with a random target style representation. 
In a recent work \cite{pizzati2020domain}, the authors perform multi-modal source-to-target image translation based on the MUNIT architecture \cite{huang2018multimodal}. 
The original datasets are augmented with additional web-crawled data, in order to reduce the gap in terms of task-unrelated data properties between sets, while at the same time highlighting the relevant task-related visual features to be matched. 
Furthermore, the style transfer method allows for multi-modal translation, therefore multiple target styles can be transferred to a single source image, thus increasing training data diversity and, in turn, enforcing the adaptation robustness
\\
Other works \cite{zhang2018fully,wu2018dcan,wu2019ace,dundar2018domain,yang2020fda} completely avoid the computational complexity of generating high resolution images with GANs by exploiting different types of style transfer techniques.
Zhang et al. \cite{zhang2018fully} adopt traditional techniques of neural style transfer \cite{gatys2015texture,gatys2016image} to separate style (low-level feature) from image content (high-level features). 
In particular, multi-level response maps of a pre-trained CNN are exploited for image synthesis, where image style is expressed by the correlation between feature maps in the form of Gram matrices.
Alternative approaches \cite{choi2019self,wu2019ace} opt for the re-normalization of source feature maps, so that their first and second order statistics match those of the target ones, by means of the AdaIN module \cite{huang2017arbitrary}. 
Differently, Dundar et al.\ \cite{dundar2018domain} make use of a photo-realistic style transfer algorithm for an iterative optimization by which both the segmentation network and the translation algorithm performances are constantly improved.
Finally, Yang et al.\ \cite{yang2020fda} remove domain-dependent visual attributes from source images by replacing the low-level frequency spectrum with that of target images, without affecting high-level semantic interpretability.  
They argue that this simple approach, despite not requiring any  additional learnable module, results in a remarkably robust adaptation performance when embedded in a multi-band framework that averages predictions with different degrees of spectral alteration.

%% file: sections/classifier_discrepancy.tex
\subsection{Classifier Discrepancy}
\label{sec:subsec:classifier}

As discussed in Section \ref{sec:subsec:domain}, feature-level adversarial domain adaptation in its original form entails the competition between the task feature extractor and a domain critic (the discriminator), whose supervisory action in principle should guide towards the cross-domain alignment of feature representations. Task-discriminativeness instead is granted by a source supervised task objective (i.e., the standard cross-entropy loss for semantic segmentation). 
\\
As highlighted by \cite{saito2017adversarial,saito2018maximum}, the major drawback of this primary form of adversarial adaptation lies in the lack of semantic awareness from the domain critic network. 
Even when the critic manages to grasp a clear expression of marginal distributions, thus effectively leading to a global statistical alignment, category-level joint-distributions necessarily remain unknown to the domain discriminator, as it 
is not provided with semantic labels when discriminating feature representations. 
A side effect to this semantic-unaware 
adaptation is that features can be placed close to class boundaries, increasing the chances of incorrect classification. Furthermore, target representations may be incorrectly transferred to a semantic category different from the actual one in the domain invariant adapted space (negative transfer), as decision boundaries are ignored in the adaptation process. 

To overcome these issues, Saito et al.\ \cite{saito2017adversarial} propose an Adversarial Dropout Regularization (ADR) approach for UDA to provide cross-domain feature alignment away from decision boundaries. 
To do so, they completely revisit the original domain adversarial scheme, by providing the task-specific dense classifier (i.e., the encoder) with a discriminative role. In particular, by means of dropout, the classifier is perturbed in order to get two distinct predictions over the same encoder output. Since the prediction variability is subject to an inverse relationship with the proximity to decision boundaries, the feature extractor is forced to produce representations far from those boundaries by minimizing the discrepancy of the two output probability maps. At the same time, the classifier has to maximize its output variation, in order to boost its capability to detect less-adapted features. 
In this redesigned adversarial scheme, the dense classifier is trained to be sensitive to semantic variations of target features, as to capture all the information stored in its neurons, which in turn are encouraged to be as diverse as possible from each other by the adversarial dropout maximization. 
On the other hand, the encoder is focused on providing categorical certainty to extracted target features, since removing task-unrelated cues weakens the possibility to achieve dissimilar predictions from the same latent representations. 
%
\\
Following the same principle of Adversarial Dropout Regularization, other approaches resort to adaptation techniques based on classifier discrepancy to achieve a semantically consistent alignment \cite{watanabe2018multichannel, qin2019generatively, saito2018maximum, luo2019taking, lee2019drop}. Saito et al.\ \cite{saito2018maximum} improve the framework in \cite{saito2017adversarial} by modifying the way of accessing multiple predictions over the same latent space. 
In place of dropout on classifier's weights, they introduce a couple of separate decoders which are simultaneously trained with source supervision, while being forced to produce dissimilar predictions by the maximization of a discrepancy loss. 
The objective is to avoid the noise sensitivity acquired by the single classifier in ADR, which is essential for the individual decoder to capture the proximity to the support of target samples, but requires an additional training stage to correctly learn the segmentation model as a whole. 
\\
The co-training strategy of exploiting a couple of distinct classifiers to infer the degree of target adaptation is further merged to the more traditional generator-discriminator adversarial framework by Luo et al.\ \cite{luo2019taking}. 
They use the discrepancy map from the two classifiers' output to weight the adversarial objective. Thereby semantic inconsistent regions highlighted by strong prediction variability get a major focus in the objective, as they should suffer from a more prominent domain shift. 
Additionally, Lee et al.\ \cite{lee2019drop} re-propose a form of adversarial dropout to get divergent predictions from a single classifier. However, they drop the adversarial scheme for a non-stochastic virtual dropout mechanism, to discover minimum distance adversarial dropout masks that maximize prediction discrepancy. In the end, they resort to a single unified objective, for a combined optimization of the encoder and decoder to align features between domains, while progressively pushing dense regions and decision boundaries far away from each other.


%% file: sections/self_training.tex
\subsection{Self-Training}
\label{sec:subsec:selftraining}

The self-training strategy entails using highly confident network predictions inferred on unlabeled data to generate pseudo-labels, to be used, in turn, to reinforce the training of the predictor with the self-taught supervision. 
This approach has been commonly employed in semi-supervised learning (SSL) \cite{grandvalet2005semi} to exploit additional unlabeled data in order to improve the prediction accuracy. 
Recently, self-training techniques have been extended to address unsupervised domain adaptation, since UDA can be considered as a variant of the SSL task, even though the additional complexity of UDA from the statistical shift of the unlabeled target data must be further taken into account. 
Indeed, concurrently learning from source annotations and target pseudo-labels implicitly promotes feature-level cross-domain alignment, while still retaining the task specificity. 
On the contrary, lacking a unified loss, other adaptation approaches, as the most successful adversarial ones, have to take care of the task-relatedness with additional training objectives. 
The critical point is that this strategy is self-referential, so careful arrangements must be adopted to avoid catastrophic error propagation. Self-training, in fact, naturally promotes more confident predictions, as the network probability output is encouraged to reach a peaked distribution (at the limit a Dirac distribution) close to the one-hot pseudo-labels. Since no form of external supervision is available on unlabeled target data, the network could yield over-confident predictions by wrongly classifying uncertain pixels. In turn, the iterative self-teaching strategy enforces prediction mistakes, through a propagation mechanism that makes the output progressively deviating from the correct solution. 
For this reason, the majority of self-training based adaptation approaches rely on various forms of pseudo-label filtering, to allow self-learning only from top confident target predictions, which are implicitly assumed to have a higher chance of being correct. 

A first class of adaptation solutions based on self-training \cite{zou2018unsupervised,zou2019confidence,li2019bidirectional} employs offline techniques for pseudo-label computation: at every update step a confidence threshold is computed by looking at the entire training set. Target segmentation maps are then directly filtered according to some confidence-based thresholding policy and used in combination with original source annotated data for the supervised learning of the segmentation network.  
\\
On this regard, Zou et al.\ \cite{zou2018unsupervised} propose one of the first UDA techniques based on self-training. 
They devise an iterative self-training optimization scheme, which alternates steps of segmentation network training on both source original and target artificial supervision and target pseudo-label estimation. 
In particular, the target pseudo-labels are treated as discrete latent variables to be computed through the minimization of a unified training objective. 
In addition, motivated by the fact that class-unaware pseudo-labels confidence filtering is intrinsically biased towards the easy (i.e., more confident) classes, they devise a class-balancing strategy by setting category-wise confidence thresholds. This should promote inter-class balance, as the same amount of top confident pixels are considered for each class, thus resulting in class-wise uniform contributions to the learning process. 
Finally, since source and target domains are supposed to share high-level scene layout, they also utilize spatial priors from source label statistics, which are inferred for each semantic category and incorporated in the training objective. 
More recently, Zou et al.\ \cite{zou2019confidence} revisit their previous work in \cite{zou2018unsupervised} by extending the pseudo-label space from one-hot maps to a continuous space defined by a probability simplex. 
In this way, by avoiding clear-cut overconfident self-supervision in the whole input image, the effect of the inherent misleading incorrect pixel predictions should be effectively reduced. 
A continuous pseudo-label space further allows them to introduce a confidence regularizing term in the training objective targeting both pseudo-label (treated as latent variables) and network weights, with the purpose of achieving output smoothness in place of sparse segmentation maps. 

In order to avoid slow offline dataset-wise processing, Pizzati et al.\ \cite{pizzati2020domain} introduce self-training with weighted pseudo-labels. A learnable confidence threshold is employed for both pseudo-label refinement and weighting, thus making pseudo-labels belonging to a continuous space, while concurrently balancing the impact of uncertain pixels. Target weighted self-generated labels are computed over a single batch, but still retaining a global view, since the confidence threshold is learned throughout the entire training phase. 

A different group of researches \cite{biasetton2019unsupervised,michieli2020adversarial,spadotto2020unsupervised} construct a self-training strategy on top of an adversarial discriminative adaptation module applied over the segmentation network output. 
In particular, on the belief that the fully convolutional discriminator 
can be regarded as performing a measure of reliability of network estimations, they exploit the discriminator output to identify reliable target predictions, which are then preserved in the pseudo-label filtering operation.  
Michieli et al.\ \cite{michieli2020adversarial} further improve the pseudo-label selection mechanisms by a region growing strategy. 
Moreover, Spadotto et al.\ \cite{spadotto2020unsupervised} propose to adopt a class-wise adaptive thresholding approach. They select the same fraction of highly confident target pixels for each semantic class, by looking at the batch-wise distribution of the discriminator probability output. In doing so, they provide the adaptation framework with both inter-class confidence flexibility and time adaptability over the training phase.

Another line of works \cite{choi2019self,chen2019domain,yang2020fda,zhou2020uncertainty} utilize various form of prediction ensembling to yield more reliable predictions over target data, on top of which pseudo-labelling is performed. 
Chen et al.\ \cite{chen2019domain} enhance the adaptation of low-level features by introducing an additional ASPP dense classification module. Hence, self-produced guidance in the form of pseudo-labels from the combined knowledge of low and high level target predictions is exploited as additional training objective. 
Yang et al.\ \cite{yang2020fda} train multiple instances of the segmentation network with multi-band spectrum adaptation to obtain distinct semantic predictors. Then, target pseudo-labels are generated from the mean prediction of the different segmenter instances, resulting in a more robust adaptation when dealing with multiple rounds of self-training.

Rather than operating directly on the predictor output, other self-training approaches \cite{choi2019self,zhou2020uncertainty} resort to an additional network to produce self-guidance over the unlabeled samples. 
Choi et al.\ \cite{choi2019self} propose a self-ensembling adaptation technique, by which a teacher network derived from student network's weights average yields predictions the student network is compelled to follow. In other words, an auxiliary predictor (the teacher network) is providing a sort of pseudo-labels, which are then used to transfer reliable knowledge to the actual predictor (the student network) by supervised training on target data. 
With regularization purposes, Gaussian noise is additionally injected on input target images and dropout weight perturbation is applied to the segmentation network to improve adaptation robustness, as student-teacher prediction consistency is enforced even under different random disturbance. 
Recently, the student-teacher self-ensembling adaptation approach is extended by Zhou et al.\ \cite{zhou2020uncertainty}, with the introduction of an uncertainty module that filters out unreliable teacher predictions by looking at self-information maps.


%% file: sections/entropy_minimization.tex
\subsection{Entropy Minimization}
\label{sec:subsec:entropy}

As already pointed out semi-supervised learning and unsupervised domain adaptation are closely related tasks: 
indeed, once source and target distributions are matched, the UDA task merely scales down to learning from an unlabeled subset of the training data. 
Therefore, it is natural that SSL approaches may inspire domain adaptation strategies, as discussed for self-training (section \ref{sec:subsec:selftraining}). 
Among the successful techniques used to address semi-supervised learning, entropy minimization has been recently introduced to UDA \cite{vu2019advent}. 
The principle behind minimizing target entropy to perform domain adaptation follows the observation that source predictions are likely to show more confidence, which in turn translates into high entropy probability outputs.
On the contrary, the segmentation network is likely to display a more uncertain behavior on target-distributed samples, as target prediction entropy maps happen to be overall quite unstable, typically being the noise pattern not confined just to the semantic boundaries. 
Thus, forcing the segmentation network to mimic the over-confident source behavior when applied to the target domain too, should effectively reduce the accuracy gap between domains. 
In other words, entropy minimization aims at penalizing classification boundaries in the latent space crossing high density regions, while jointly encouraging well-clustered target representations properly sorted out by decision boundaries. 

In its simplest fashion \cite{vu2019advent} entropy minimization is performed at a pixel-level, so that each spatial unit of the prediction map brings an independent contribution to the final objective. However, the basic approach suffers from some intrinsic limitations, demanding further arrangements to boost the adaptation performance \cite{vu2019advent,chen2019domain,yang2020fda}. 
To leverage structural information of semantic maps, Vu et al.\ \cite{vu2019advent} propose a global adversarial optimization to enforce distribution alignment over source and target entropy maps. In doing so, they rely on a domain discriminator to capture global patterns differentiating samples from separate domains, thus achieving a more semantically meaningful cross-domain match of entropy behavior. Class-wise priors on label distributions inferred from source annotations are further enforced on target predictions to avoid class imbalance towards easy classes.
\\
In a following publication, Chen et al.\ \cite{chen2019domain} observe that the entropy minimization objective can be seriously hindered by the gradient predominance of more confident predictions. Indeed, moving from high to low uncertainty areas, the gradient rapidly increases, and its value tends to infinity as the output probability distribution tends to the delta function. This probability imbalance in general prevents the segmentation network from learning over areas with little accuracy, whose gradients result much lower that those of easy-to-transfer image regions. 
To address this issue, they devise a maximum squares loss, which produces a gradient signal that grows linearly with the input probability. 
They also face class unbalance by introducing a category-wise weighting factor based on target distribution from prediction maps in place of source annotations, as they argue that source class statistics may significantly deviate from target ones. 
\\
Very recently, Yang et al.\ \cite{yang2020fda} have added an entropy minimization technique as an additional module to their adaptation scheme. 
The intent is to seek a regularization effect over the training on unlabeled target data, accomplished by pushing the decision boundaries away from high-density regions in the target latent space, with basically no overhead to the actual framework. 
The strength of the approach is enhanced by the combined application of other adaptation modules to achieve domain alignment. This, in fact, shifts the UDA task towards SSL, thus making entropy minimization more effective. 
Moreover, to avoid excessive emphasis on low entropy predictions, they adopt a penalty function that increase the focus on less-adapted high entropy regions of target images.



%% file: sections/adaptation_newresearchdirections.tex
\subsection{New Research Directions}
\label{sec:subsec:newresearch}
Unsupervised Domain Adaptation in its original interpretation aims at addressing the domain shift by transferring representations concerning a specific and well-defined set of semantic categories shared across source and target data. 
This follows the assumption that the target domain contains only instances of the classes which can be found in source samples. 
Nevertheless, while being a reasonable hypothesis that does not hinder the generality of the adaptation task,
in practice it is common that images from a novel domain may contain objects from unseen categories.
\\  
Moving in the direction of a more general definition of the adaptation objective,  
some works have tackled the open-set domain adaptation \cite{busto2017open} applied to the image classification task, which entails unknown categories peculiar to the target domain not present in the source one, but they still retain a somewhat strict prior definition of the adaptation settings class-wise. 
Recently, a few novel approaches \cite{saito2020universal,zhuo2019unsupervised} have proposed to relax the common premises on the domain adaptation settings, to effectively move towards a more realistic scenario where little can be inferred a priori about target data properties, thus widening the applicability to real-world solutions. 
\\
Saito et al.\ \cite{saito2020universal}, for instance, introduce the Universal Domain Adaptation problem, allowing for basically no beforehand characterization of target classes. In particular, they resort to a neighborhood clustering technique to assign each target sample to either a source class or to the unknown category without any supervision. Then, the matching of cross-domain representations is enforced by entropy minimization to achieve domain alignment. 
\\
A step further is attained by solving the recognition of unseen target categories, which have to be individually learnt rather than simply acknowledged as unknown. 
Zhuo et al.\ \cite{zhuo2019unsupervised} address what they call the Unsupervised Open Domain Recognition task, where the objective is to learn to correctly classify target samples of unknown classes. To do so, they reduce the domain shift between source and target sets by an instance matching discrepancy minimization, weighted according to feature similarity. Once the semantic predictor has achieved domain invariance, classification knowledge can be safely transferred from known to unknown categories by a graph CNN module.

Despite the aforementioned adaptation approaches have proven to be quite effective for the image classification task, further adjustments needs to be done to deal with the additional complexity of feature representations of a semantic segmentation network.
On this regard, a novel Boundless Unsupervised Domain Adaptation (BUDA) task is proposed by Bucher et al.\ \cite{bucher2020buda} specifically for semantic segmentation. Similarly to UODR \cite{zhuo2019unsupervised}, the standard domain adaptation problem is \textit{unbounded} to explicitly handle instances of new unseen target classes, while relying solely on a minimal semantic prior in the form of class names, that are supposed to be known in advance. 
Thus, the overall task is decoupled in the domain adaptation and zero-shot learning problems. First, the domain adaptation of categories in common between source and target domains is performed, via an entropy minimization technique carefully designed to avoid incorrect alignment over unseen target classes. Then, a zero-shot learning strategy \cite{bucher2019zero} is exploited to transfer knowledge from seen to unseen classes, by a generative model able to synthesize visual features conditioned by class descriptors.

Another closely related research direction which is currently gaining wider and wider interest among the research community, is the continual learning task. Continual learning could be regarded as a particular case of transfer learning, where the data domain distribution changes at every incremental step and the models should perform well on all the domain distributions. For instance, in class-incremental learning, the learned model is updated to perform a new task whilst preserving previous capability. Initially proposed for image classification \cite{kirkpatrick2017overcoming,li2017learning} and object detection \cite{shmelkov2017incremental} it has been recently explored also for semantic segmentation \cite{michieli2019incremental,michieli2019knowledge,cermelli2020modeling}. 
Another formulation of this problem regards the coarse-to-fine refinement at the semantic level, in which previous knowledge acquired on a coarser task is exploited to perform a finer task, hence modifying the labels distribution \cite{mel2020incremental}.

%% file: sections/datasets.tex
\section{A Case Study: Synthetic to Real Adaptation for Semantic Understanding of Road Scenes}
\label{sec:datasets}

The main aspect for UDA techniques is the ability to transfer knowledge acquired on one dataset to a different context. Hence, the considered data play a fundamental role in the design and evaluation of UDA algorithms. In this section, we focus on one of the most interesting application scenarios: i.e., the ability to transfer knowledge acquired on synthetic datasets (source domain), where labels are relatively inexpensive and can be easily produced with computer graphics engines, to real world ones (target domain), where annotations are highly expensive, time-consuming and error-prone. 
Many of the works dealing with this task focus on urban scenes mainly for four reasons: 
\begin{enumerate}
\item Autonomous driving is nowadays one of the biggest research areas  and massive fundings support this research \cite{smith2019gartner};
\item Many synthetic and real world datasets are publicly available for this scenario  \cite{Richter2016,ros2016,cordts2016cityscapes,neuhold2017mapillary};
\item Autonomous vehicles should fully understand the surrounding environment  to plan decisions \cite{michieli2018} and such navigation task in the environment could be encountered in many other applications, e.g., in the robotics field;
\item The first works on the topic addressed this setting and it has become the de-facto standard for performance  comparison  with the state-of-the-art in the UDA for semantic segmentation field.
\end{enumerate}



Before presenting the more commonly used synthetic and real world datasets, we stress that in the unsupervised domain adaptation scenario the expensive labels of real samples in the target domain are not needed for training. However, a limited number of real target samples must be manually labeled for testing (and sometimes validating) the performance of the algorithms. On the other hand, large synthetic datasets corresponding to the source domain are equipped with annotations which are exploited for supervised training.

\subsection{Source Domain: Synthetic Datasets of Urban Scenes}

One of the first large scale synthetic datasets for urban driving is the \textbf{GTA5} dataset \cite{Richter2016}. It contains $24966$  synthetic $1914 \times 1052$ $\mathrm{px}$ images with pixel-level semantic annotation. The images have been rendered using the open-world video game \textit{Grand Theft Auto V} and are all from the car perspective in the streets of American-style virtual cities (resembling the ones in California).
The images have an impressive visual quality and are very realistic since the rendering engine comes from a high budget commercial production. The data is labeled into 19 semantic classes which are compatible with the ones of real-world datasets as Cityscapes or Mapillary (after a proper re-mapping of labels). 

The \textbf{SYNTHIA-RAND-CITYSCAPES} dataset has been sampled using the same simulator as the \textbf{SYNTHIA} dataset \cite{ros2016} and contains $9400$ synthetic $1280 \times 760$ $\mathrm{px}$ images with pixel level semantic annotation.
The images have been rendered with an ad-hoc graphic engine, allowing to obtain a large variability of photo-realistic street scenes (in this case they come from virtual European-style towns in different environments under various light and weather conditions).
On the other hand, the visual quality is lower than the commercial video game GTA5. 
The semantic labels are compatible with $16$ classes of real world datasets like Cityscapes and Mapillary. For the evaluation, either $13$ or $16$ classes are taken into consideration. 

\textbf{CARLA} (CAR Learning to Act) \cite{dosovitskiy2017carla} is an open-source simulator for autonomous driving research built over the  Unreal Engine 4 rendering software. It has been designed to grant large flexibility and both the physics and the rendering simulations are quite realistic. Two virtual towns have been designed: \textit{Town 1} with $2.9$ km of drivable roads and \textit{Town 2} with $1.4$ km of drivable roads. 3D artists first laid roads and sidewalks, then they placed houses, terrain, vegetation, and traffic infrastructure to resemble a realistic environment. Then, dynamic objects, like cars and pedestrian spawn from specific coordinates. The APIs of CARLA give access to a semantic segmentation camera that can distinguish $13$ different classes. 
This feature makes possible to sample urban datasets very quickly, easily and with a large control on the variability of samples. Another relevant aspect is that anyone can create its own dataset based on the specific needs customizing the open-source simulator.

\subsection{Target Domain: Real World Datasets of Urban Scenes}

The \textbf{Cityscapes} dataset \cite{cordts2016cityscapes} contains $2975$ color images of resolution $2048 \times 1024$ $\mathrm{px}$ captured on the streets of $50$ European cities.  
The images have pixel-level semantic annotation with a total of $34$ semantic classes.
For the evaluation of UDA approaches typically the original training set (without the labels) is used for unsupervised adaptation, while the $500$ images in the original validation set are used as a test set (since the test set labels have not been made available).

The \textbf{Mapillary} dataset \cite{neuhold2017mapillary} contains  $25000$ variable (typically very high) resolution color images  taken from different devices in many different locations around the world. 
The variability in classes, appearance, acquisition settings and geo-localization makes the dataset the most complete and the one of the highest quality in the field.
The $152$ object-level semantic annotations are often re-conducted to the classes present in the Cityscapes dataset, e.g., following the mapping in \cite{kim2018attribute}. 
The training images (without the labels) are used for unsupervised adaptation and the images in the original validation set are exploited as test set.

The \textbf{Oxford RobotCar Dataset} \cite{maddern20171} contains about $1000$ km of images recorded driving in the central part of Oxford (UK). The same route (approximately $10$ km long) has been repeatedly traversed for almost a year and $20$ million images under different weather and light conditions have been collected by the $6$ cameras the car was equipped with. All data are associated also to LiDAR, GPS and INS ground truth.

In \cite{chen2017no}, \textbf{Google Street View} is used to collect a large number of not annotated street images from Rome, Rio, Tokyo and Taipei. These cities have been chosen to ensure enough visual variations and the locations in the cities have been randomly selected. 
Using the time-machine feature of Google Street View it has been possible to capture images of the same street scene at different times in order to extract static objects priors. They collected $1600$ ($647 \times 1280$ $\mathrm{px}$) image pairs at $1600$ different locations per city. A set of $100$ random images for each city have been annotated with pixel-level semantic labels for testing purposes.

\subsection{Methods Comparison}

\input{sections/tables.tex}

In this section, the main results of the approaches described in the previous chapters are summarized and briefly discussed. For the sake of brevity, only the most widely used datasets are considered in this section: namely, GTA5 and SYNTHIA as source datasets, Cityscapes and Mapillary as target datasets. The others previously introduced datasets are less commonly used even if they could  in principle be used being equipped with semantic segmentation annotations.\\
Before digging into the description of the results of existing methods we warn the reader to be aware that different evaluation protocols and experimental setups exist making the direct comparison of the final accuracy results not always faithful. For instance, differences in input image resolution, batch size, backbone network architecture and other training parameters may alter the comparison. 
As for the metrics, the per-class Intersection over the Union (IoU$_i$ for a generic class $i$) and the mean IoU (mIoU) are the most common. They  are defined as:
\begin{equation}
\mathrm{IoU}_i = \frac{TP_i}{FP_i + FN_i + TP_i}
\end{equation}
\begin{equation}
\displaystyle \mathrm{mIoU} = \sum_{i=1}^N \frac{\mathrm{IoU}_i}{N}
\end{equation}

Where $TP_i$, $FP_i$ and $FN_i$ represent respectively the number of true positive, false positive and false negative pixels for a generic class $i$, and $N$ is the number of classes.

Table~\ref{tab:GTA_CS} shows the mIoU results for different methods grouped by the employed backbone network when adapting source knowledge from GTA5 to Cityscapes. The results are grouped by backbone 
and we can appreciate that the results could greatly vary depending on the method and on the evaluation protocol used.

The entries in this table are scattered in Figure~\ref{fig:miou_papers_GC} grouped by backbone architecture to show the mIoU values and the corresponding mean (only the backbones with at least 3 entries are considered).
In general we can see that ResNet-based approaches outperform the competitors. 
Moreover, the most widely diffused architectures are ResNet-101 and VGG-16. 
Employing the ResNet-101 architecture looks to be the best option for full comparison with the existing literature.

Similarly, Table~\ref{tab:SYNTHIA_CS} reports the results grouped by backbone when adapting source knowledge from SYNTHIA to Cityscapes. The table reports two setups, i.e., considering either $13$ or $16$ classes, since both are often considered in this case.
The respective entries of mIoU$_{16}$ are scattered in Figure~\ref{fig:miou_papers_SC} to give an overview of the most widely used techniques and of the results achieved. Also in this scenario, VGG-16 is the most popular architecture followed by ResNet-101, which generally shows higher results.

Some recent works consider also the Mapillary dataset 
adapting from either GTA5 or SYNTHIA as source datasets, as before. While, in this case, the comparison is quite limited \cite{michieli2020adversarial,biasetton2019unsupervised,spadotto2020unsupervised} (currently, to the best of our knowledge, the highest performing approach is \cite{spadotto2020unsupervised}, that achieves a mIoU of $41.9$ when adapting from GTA5), we believe Mapillary represents a much more variable dataset than Cityscapes which should be taken into considerations from future works. 

%% file: sections/tables.tex
\begin{table}[tbp]
  \centering
  \caption{Mean IoU (mIoU) for different methods grouped by backbone in the scenario adapting source knowledge from GTA5 to Cityscapes. }
   \setlength{\tabcolsep}{9pt}
    \begin{tabular}{|l|c|c||l|c|c|}
    \hline
    \textbf{Method} & \textbf{Backbone} & \textbf{mIoU}  & \textbf{Method} & \textbf{Backbone} & \textbf{mIoU} \\\hline
    Biasetton et al. \cite{biasetton2019unsupervised} & ResNet-101 & 30.4  & Chen et al. \cite{chen2018road} & VGG-16 & 35.9 \\
    Chang et al. \cite{chang2019all} & ResNet-101 & 45.4  & Chen et al. \cite{chen2019crdoco} & VGG-16 & 38.1 \\
    Chen et al. \cite{chen2018road} & ResNet-101 & 39.4  & Choi et al. \cite{choi2019self} & VGG-16 & 42.5 \\
    Chen et al. \cite{chen2019domain} & ResNet-101 & 46.4  & Du et al. \cite{du2019ssf} & VGG-16 & 37.7 \\
    Du et al. \cite{du2019ssf} & ResNet-101 & 45.4  & Hoffman et al. \cite{hoffman2016fcns} & VGG-16 & 27.1 \\
    Gong et al. \cite{gong2019dlow}  & ResNet-101 & 42.3  & Hoffman et al. \cite{hoffman2017cycada} & VGG-16 & 35.4 \\
    Hoffman et al. \cite{hoffman2017cycada} & ResNet-101 & 42.7* & Huang et al. \cite{huang2018domain} & VGG-16 & 32.6 \\
    Li et al. \cite{li2019bidirectional} & ResNet-101 & 48.5  & Li et al. \cite{li2019bidirectional} & VGG-16 & 41.3 \\
    Lian et al. \cite{lian2019constructing} & ResNet-101 & 47.4  & Lian et al. \cite{lian2019constructing} & VGG-16 & 37.2 \\
    Luo et al. \cite{luo2019significance} & ResNet-101 & 42.6  & Luo et al. \cite{luo2019significance} & VGG-16 & 34.2 \\
    Luo et al. \cite{luo2019taking} & ResNet-101 & 43.2  & Luo et al. \cite{luo2019taking} & VGG-16 & 36.6 \\
    Michieli et al. \cite{michieli2020adversarial} & ResNet-101 & 33.3  & Saito et al. \cite{saito2018maximum} & VGG-16 & 28.8 \\
    Spadotto et al. \cite{spadotto2020unsupervised} & ResNet-101 & 35.1  & Sankaranarayanan et al. \cite{sankaranarayanan2018learning} & VGG-16 & 37.1 \\
    Tsai et al. \cite{tsai2018learning} & ResNet-101 & 42.4  & Tsai et al. \cite{tsai2018learning} & VGG-16 & 35.0 \\
    Tsai et al. \cite{tsai2019domain} & ResNet-101 & 46.5  & Tsai et al. \cite{tsai2019domain} & VGG-16 & 37.5 \\
    Vu et al. \cite{vu2019advent} & ResNet-101 & 45.5  & Vu et al. \cite{vu2019advent} & VGG-16 & 36.1 \\
    Wu et al. \cite{wu2018dcan} & ResNet-101 & 38.5  & Wu et al. \cite{wu2018dcan} & VGG-16 & 36.2 \\
    Yang et al. \cite{yang2020fda} & ResNet-101 & 50.5  & Yang et al. \cite{yang2020fda} & VGG-16 & 42.2 \\
    Zhang et al. \cite{zhang2018fully} & ResNet-101 & 47.8  & Zhang et al. \cite{zhang2017curriculum} & VGG-16 & 28.9 \\
    Zou et al. \cite{zou2019confidence} & ResNet-101 & 47.1  & Zhang et al. \cite{zhang2019curriculum} & VGG-16 & 31.4 \\\cdashline{1-3}
    Murez et al. \cite{murez2018image} & ResNet-34 & 31.8  & Zhou et al. \cite{zhou2020uncertainty} & VGG-16 & 47.8 \\\cdashline{1-3}
    Lian et al. \cite{lian2019constructing} & ResNet-38 & 48.0  & Zhu et al. \cite{zhu2018penalizing} & VGG-16 & 38.1* \\
    Zou et al. \cite{zou2018unsupervised} & ResNet-38 & 47.0  & Zou et al. \cite{zou2018unsupervised} & VGG-16 & 36.1 \\\cdashline{4-6}
    Zou et al. \cite{zou2019confidence} & ResNet-38 & 49.8  & Hong et al. \cite{hong2018conditional} & VGG-19 & 44.5 \\\cdashline{1-6}
    Lee et al. \cite{lee2019drop} & ResNet-50 & 35.8  & Chen et al. \cite{chen2019crdoco} & DRN-26 & 45.1 \\
    Saito et al. \cite{saito2017adversarial} & ResNet-50 & 33.3  & Dundar et al. \cite{dundar2018domain} & DRN-26 & 38.3 \\
    Wu et al. \cite{wu2018dcan} & ResNet-50 & 41.7  & Hoffman et al. \cite{hoffman2017cycada} & DRN-26 & 39.5 \\\cdashline{1-3}
    Hoffman et al. \cite{hoffman2017cycada} & MobileNet-v2 & 37.3* & Huang et al. \cite{huang2018domain} & DRN-26 & 40.2 \\
    Toldo et al. \cite{toldo2020unsupervised} & MobileNet-v2 & 41.1  & Liu et al. \cite{liu2017unsupervised} & DRN-26 & 39.1* \\
    Zhu et al. \cite{zhu2017unpaired} & MobileNet-v2 & 29.3* & Yang et al. \cite{yang2020phase} & DRN-26 & 42.6 \\\cdashline{1-3}
    Murez et al. \cite{murez2018image} & DenseNet & 35.7  & Zhu et al. \cite{zhu2017unpaired} & DRN-26 & 39.6* \\\cdashline{1-6}
    Huang et al. \cite{huang2018domain} & ERFNet & 31.3  & Saito et al. \cite{saito2018maximum} & DRN-105 & 39.7 \\\hline
    \end{tabular}%
 \begin{flushright}    *: values from results of competing works.\end{flushright}
  \label{tab:GTA_CS}%
\end{table}%

\begin{figure}[tbp]
\includegraphics[width=\textwidth]{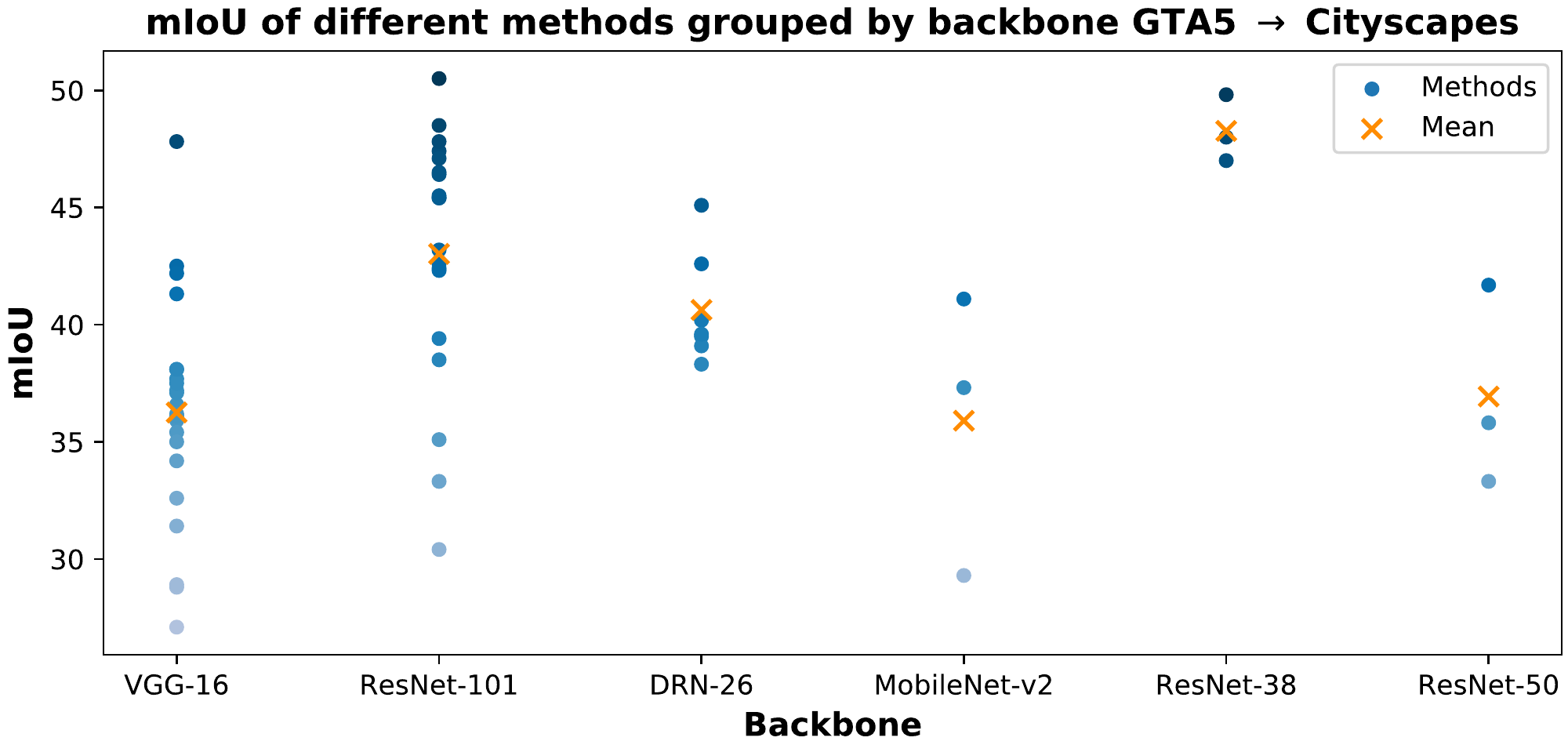}
\caption{Mean IoU (mIoU) of different methods grouped by backbone in the scenario adpating source knowledge from GTA5 to Cityscapes (see Table~\ref{tab:GTA_CS}). Backbones are sorted by decreasing number of entries. Orange crosses represent the per-backbone mean mIoU. Only the backbones with $3$ or more entries were displayed.}
\label{fig:miou_papers_GC}
\end{figure}

\begin{table}[tbp]
  \centering
  \setlength{\tabcolsep}{4pt}
  \caption{Mean IoU (mIoU) for different methods grouped by backbone in the scenario adapting source knowledge from SYNTHIA to Cityscapes. The table reports the mIoU computed over $13$ or $16$ semantic classes depending on the label set used in the corresponding paper. }
    \begin{tabular}{|l|c|c|c||l|c|c|c|}
    \hline
    \textbf{Method} & \textbf{Backbone} & \textbf{mIoU$_{13}$} & \textbf{mIoU$_{16}$} & \textbf{Method} & \textbf{Backbone} & \textbf{mIoU$_{13}$} & \textbf{mIoU$_{16}$} \\\hline
    Biasetton et al. \cite{biasetton2019unsupervised} & ResNet-101 & -     & 30.2  & Chen et al. \cite{chen2017no}  & VGG-16 & 35.7  & - \\
    Bucher et al. \cite{bucher2020buda}  & ResNet-101 & -     & 36.2  & Chen et al. \cite{chen2018road} & VGG-16 & -     & 36.2 \\
    Chang et al. \cite{chang2019all} & ResNet-101 & -     & 41.5  & Chen et al. \cite{chen2018road} & VGG-16 & 41.8* & 36.2* \\
    Chen et al. \cite{chen2019domain} & ResNet-101 & 48.2  & 41.4  & Chen et al. \cite{chen2019crdoco} & VGG-16 & -     & 38.2 \\
    Du et al. \cite{du2019ssf} & ResNet-101 & 50.0  & -     & Chen et al. \cite{chen2019learning} & VGG-16 & 43.0  & 37.3 \\
    Li et al. \cite{li2019bidirectional} & ResNet-101 & 51.4  & -     & Choi et al. \cite{choi2019self}  & VGG-16 & 46.6  & 38.5 \\
    Lian et al. \cite{lian2019constructing} & ResNet-101 & 53.3  & 46.7  & Du et al. \cite{du2019ssf} & VGG-16 & 43.4  & - \\
    Luo et al. \cite{luo2019significance} & ResNet-101 & 46.3  & -     & Hoffman et al. \cite{hoffman2016fcns} & VGG-16 & 17.0  & 20.2*  \\
    Luo et al. \cite{luo2019taking} & ResNet-101 & 47.8  & -     & Huang et al. \cite{huang2018domain} & VGG-16 & -     & 30.7* \\
    Michieli et al. \cite{michieli2020adversarial} & ResNet-101 & -     & 31.3  & Lee et al. \cite{lee2018spigan} & VGG-16 & 42.4* & 36.8 \\
    Spadotto et al. \cite{spadotto2020unsupervised} & ResNet-101 & -     & 34.6  & Li et al. \cite{li2019bidirectional} & VGG-16 & -     & 39.0  \\
    Tsai et al \cite{tsai2019domain} & ResNet-101 & 46.5  & 40.0  & Lian et al. \cite{lian2019constructing} & VGG-16 & 42.6  & 35.9  \\
    Tsai et al. \cite{tsai2018learning} & ResNet-101 & 46.7  & -     & Luo et al. \cite{luo2019taking} & VGG-16 & 39.3  & - \\
    Vu et al. \cite{vu2019advent} & ResNet-101 & 48.0  & 41.2  & Luo et al. \cite{luo2019significance} & VGG-16 & 37.2  & - \\
    Vu et al. \cite{vu2019dada} & ResNet-101 & 49.8  & 42.6  & Sankaran. et al. \cite{sankaranarayanan2018learning} & VGG-16 & 42.1* & 36.1 \\
    Wu et al. \cite{wu2018dcan} & ResNet-101 & -     & 36.5  & Tsai et al \cite{tsai2019domain} & VGG-16 & 39.6  & 33.7   \\
    Yang et al. \cite{yang2020fda} & ResNet-101 & 52.5  & -     & Tsai et al. \cite{tsai2018learning} & VGG-16 & 37.6  & - \\
    Zou et al. \cite{zou2019confidence} & ResNet-101 & 50.1  & 43.8  & Vu et al. \cite{vu2019advent} & VGG-16 & 36.6  & 31.4 \\\cdashline{1-4}
    Zou et al. \cite{zou2018unsupervised} & ResNet-38 & -     & 38.4  & Wu et al. \cite{wu2018dcan} & VGG-16 & -     & 35.4 \\\cdashline{1-4}
    Wu et al. \cite{wu2018dcan} & ResNet-50 & 48.4  & 42.5  & Yang et al. \cite{yang2020fda} & VGG-16 & -     & 40.5 \\\cdashline{1-4}
    Hoffman et al \cite{hoffman2017cycada} & MobileNet-v2 & -     & 27.5* & Yang et al. \cite{yang2020phase} & VGG-16 & 48.7  & 41.1 \\
    Toldo et al. \cite{toldo2020unsupervised} & MobileNet-v2 & -     & 32.6  & Zhang et al. \cite{zhang2017curriculum} & VGG-16 & 34.8* & 29.0 \\
    Zhu et al. \cite{zhu2017unpaired} & MobileNet-v2 & -     & 24.2* & Zhang et al. \cite{zhang2019curriculum} & VGG-16 & -     & 29.7 \\\cdashline{1-4}
    Chen et al. \cite{chen2019crdoco} & DRN-26 & -     & 33.4  & Zhou et al. \cite{zhou2020uncertainty} & VGG-16 &  48.6  & 41.5 \\
    Dundar et al. \cite{dundar2018domain} & DRN-26 & -     & 29.5  & Zhu et al. \cite{zhu2018penalizing} & VGG-16 & 40.3* & 34.2* \\
    Liu et al. \cite{liu2017unsupervised} & DRN-26 & -     & 28.0* & Zou et al. \cite{zou2018unsupervised} & VGG-16 & 36.1  & 35.4 \\\cdashline{5-8}
    Zhu et al. \cite{zhu2017unpaired} & DRN-26 & -     & 27.1* & Hong et al. \cite{hong2018conditional} & VGG-19 & -     & 41.2 \\\cdashline{1-4} 
    Saito et al. \cite{saito2018maximum} & DRN-105 &  43.5* & 37.3* & \multicolumn{4}{r|}{ \textit{*: values 
    from results  of  competing works.}}   \\\hline
    \end{tabular}%
  \label{tab:SYNTHIA_CS}%
\end{table}%

\begin{figure}[tbp]
\includegraphics[width=\textwidth]{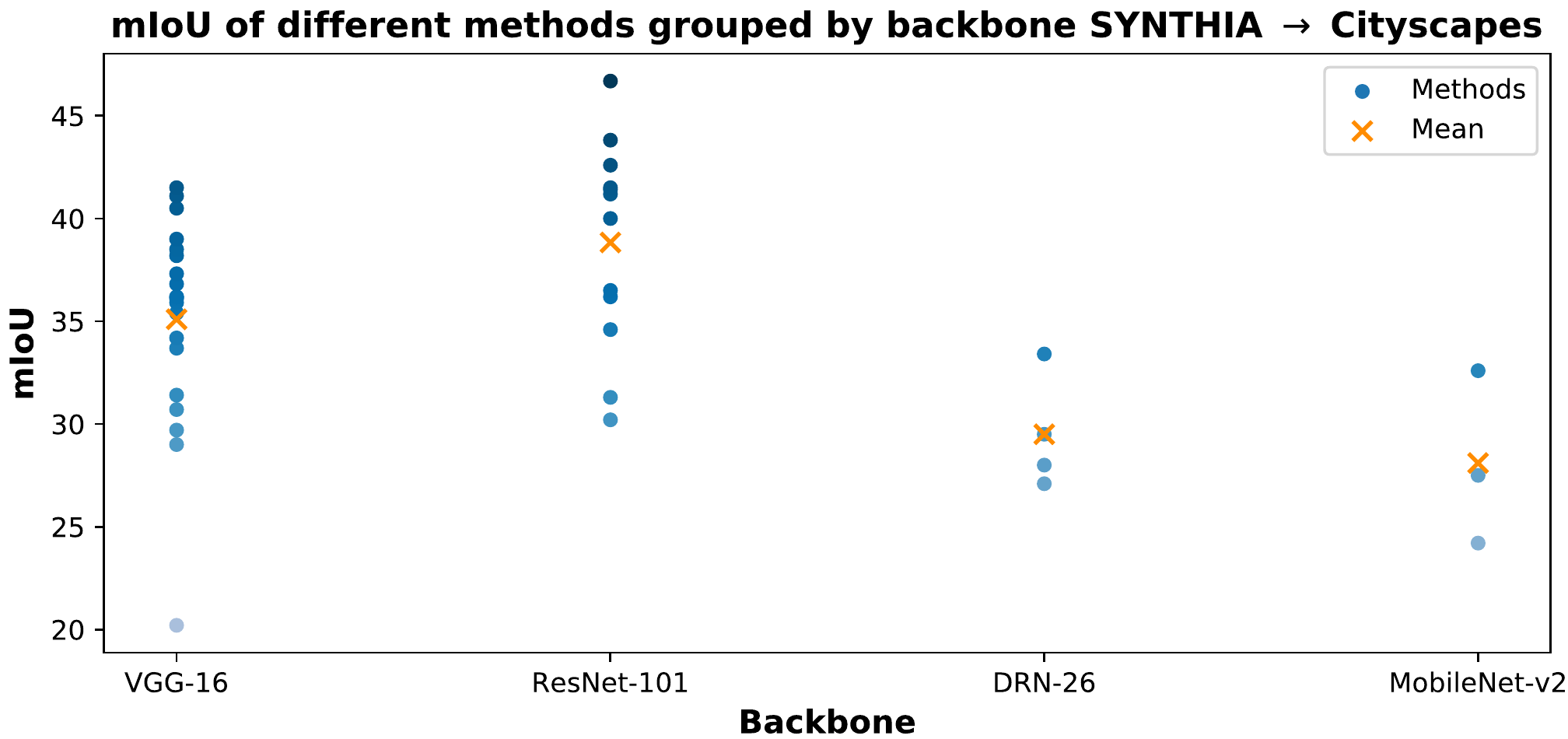}
\caption{Mean IoU on $16$ classes (mIoU$_{16}$) of different methods grouped by backbone in the scenario adpating source knowledge from SYNTHIA to Cityscapes (see Table~\ref{tab:SYNTHIA_CS}). Backbones are sorted by decreasing number of entries. Orange crosses represent the per-backbone mean mIoU. Only the backbones with $3$ or more entries were displayed.}
\label{fig:miou_papers_SC}
\end{figure}


%% file: sections/conclusion.tex
\section{Conclusion and Future Directions}
\label{sec:conclusion}

In this paper, we gave a comprehensive overview of the recent advancements in Unsupervised Domain Adaptation for semantic segmentation. This is a very relevant task since deep learning architectures for semantic segmentation require huge amount of labeled training samples, which in many practical settings are not available due to the complex labeling procedure. For this reason, a wide range of different UDA approaches for this task have been proposed in the recent years exploiting many ideas.

In order to organize the wide range of existing approaches we started from grouping them at a high-level, based on where the domain adaptation is performed: namely, at input-level (i.e., on the images provided to the network), at feature-level, at output-level or at some ad-hoc network levels. After this macroscopic subdivision, we moved to the actual review of the literature  in the field, dividing the existing works in $7$ (non mutually exclusive) categories: i.e., based on adversarial learning, on generative approaches, on the analysis of the classifier discrepancies, on self-training, on entropy minimization, on curriculum learning and finally on multi-task learning. 
For each category, we presented the most successful approaches and we carefully summarized the main ideas of each contribution.
Finally, we discussed a case study: the synthetic to real adaptation for semantic understanding of road scenes. Besides being a very relevant task since it is one of the key enabling technologies for autonomous driving, it has also being used for the evaluation of many papers in the field and we concluded the survey comparing the accuracy of many different works grouped by backbone architecture on this task.

We believe that UDA for semantic segmentation is an open research field with large room for improvements, as proved by the fact that  even  the best approaches have performance still far from the ones of supervised training on the target dataset. More refined and performing approaches based on the various schemes presented in the paper are continuously appearing. Further novel research directions will regard open problems in the field, as for example open-set and boundless-set UDA in the semantic segmentation task. Additionally, knowledge acquired for UDA may be beneficial for other closely related tasks as continual learning, where data distribution changes multiple times.
